\documentclass[pdflatex,sn-nature]{sn-jnl}

\usepackage{graphicx}%
\usepackage{multirow}%
\usepackage{amsmath,amssymb,amsfonts}%
\usepackage{amsthm}%
\usepackage{mathrsfs}%
\usepackage{xcolor}%
\usepackage{textcomp}%
\usepackage{manyfoot}%
\usepackage{booktabs}%
\usepackage{algorithm}%
\usepackage{algorithmicx}%
\usepackage{algpseudocode}%
\usepackage{listings}%
\usepackage{makecell}
\usepackage{tabularx}
\usepackage{array}
\usepackage{xr}

\theoremstyle{thmstyleone}%
\theoremstyle{thmstyletwo}%

\theoremstyle{thmstylethree}%

\begin{document}

\title[Tabular foundation models for in-context prediction of molecular properties]{Tabular foundation models for in-context prediction of molecular properties} 

\author[1]{\fnm{Karim K.} \sur{Ben Hicham}}

\author[1]{\fnm{Jan G.} \sur{Rittig}}

\author[2]{\fnm{Martin} \sur{Grohe}}

\author*[4,1,3]{\fnm{Alexander} \sur{Mitsos}}\email{amitsos@alum.mit.edu}

\affil[1]{\orgname{RWTH Aachen University, Process Systems Engineering (AVT.SVT)}, 
    \orgaddress{\city{Aachen}, \country{Germany}}}

\affil[2]{\orgname{Lehrstuhl Informatik 7, RWTH Aachen University}, 
    \orgaddress{\city{Aachen}, \country{Germany}}}

\affil[3]{\orgname{Forschungszentrum J\"ulich GmbH, Institute of Climate and Energy Systems ICE-1: Energy Systems Engineering}, 
    \orgaddress{\city{J\"ulich}, \country{Germany}}}

\affil[4]{\orgname{JARA-CSD}, 
    \orgaddress{\city{Aachen}, \country{Germany}}}

\abstract{
\unboldmath
Accurate molecular property prediction is central to drug discovery, catalysis, and process design, yet real-world applications are often limited by small datasets.
Molecular foundation models provide a promising direction by learning transferable molecular representations; however, they typically involve task-specific fine-tuning, require machine learning expertise, and often fail to outperform classical baselines.
Tabular foundation models (TFMs) offer a fundamentally different paradigm: they perform predictions through in-context learning, enabling inference without task-specific training.
Here, we evaluate TFMs in the low- to medium-data regime across both standardized pharmaceutical benchmarks and chemical engineering datasets.
We evaluate both frozen molecular foundation model representations, as well as classical descriptors and fingerprints.
Across the benchmarks, the approach shows excellent predictive performance while reducing computational cost, compared to fine-tuning, with these advantages also transferring to practical engineering data settings.
In particular, combining TFMs with CheMeleon embeddings yields up to 100\% win rates on 30 MoleculeACE tasks, while compact RDKit2d and Mordred descriptors provide strong descriptor-based alternatives.
Molecular representation emerges as a key determinant in TFM performance, with molecular foundation model embeddings and 2D descriptor sets both providing substantial gains over classic molecular fingerprints on many tasks.
These results suggest that in-context learning with TFMs provides a highly accurate and cost-efficient alternative for property prediction in practical applications.
}

\maketitle

\section{Introduction}
The ability to predict molecular properties is essential for data-driven decision-making in product, process, and catalyst design,
where reliable estimates of biological, physicochemical, or quantum-mechanical properties are needed to prioritize candidates before costly experiments~\cite{yuDeepLearningLargescale2024,eraqiMolecularPropertyPrediction2025,rittig2025molecular}.
In practice, however, many relevant prediction problems are defined by small- to medium-sized datasets rather than by the large-scale regimes in which deep learning has achieved its greatest successes~\cite{altae-tranLowDataDrug2017,eraqiMolecularPropertyPrediction2025, tetkoBeAwareOverfitting2024}.
This data-limited setting remains one of the main challenges in molecular machine learning.

Molecular foundation models have been proposed, in part, to address this challenge of low data availability: by pretraining on large, relatively inexpensive collections of unlabeled molecular strings, structures, or related datasets, they aim to learn transferable chemical representations that can support accurate prediction on heterogeneous downstream tasks with only limited labeled data~\cite{praskiBenchmarkingPretrainedMolecular2025,rossLargescaleChemicalLanguage2022,wangMolecularContrastiveLearning2022,burnsDeepLearningFoundation2026}.
Yet in practice, their downstream use depends on task-specific fine-tuning, which is prone to overfitting~\cite{kumarFineTuningCanDistort2022}, can be sensitive to hyperparameter choices, and still often fails to consistently outperform strong classical baselines such as random forests or gradient boosting tree models trained on fixed molecular fingerprints~\cite{grinsztajnWhyTreebasedModels2022b, vantilborgExposingLimitationsMolecular2022,burnsDeepLearningFoundation2026}.
The problem is therefore not only how to pretrain informative molecular foundation models, but how to harness the information they contain robustly and efficiently for small downstream datasets.
Another, arguably more practical, limiting factor for broad real-world adoption is the required ML expertise and computational budget.

In this context, recent advances in tabular foundation models (TFMs), such as TabPFN~\cite{hollmannAccuratePredictionsSmall2025,hollmannTabPFNTransformerThat2023} and TabICL~\cite{quTabICLv2BetterFaster2026, quTabICLTabularFoundation2025}, are highly promising.
TFMs are pretrained on a large variety of purely synthetic datasets generated via structural causal models (SCMs) and then perform in-context learning: training-free, direct prediction of missing labels in newly provided tabular datasets, enabling inference without task-specific fine-tuning.
These models have achieved state-of-the-art performance on comprehensive benchmarks of small- to medium-sized tabular datasets~\cite{ericksonTabArenaLivingBenchmark2025}, and exhibit substantial generalization capabilities~\cite{hollmannAccuratePredictionsSmall2025, quTabICLv2BetterFaster2026,grinsztajn2025tabpfn}.

This improved performance-per-compute trade-off suggests a promising alternative to conventional fine-tuning workflows in small- to medium-sized molecular datasets.
Molecular representations can be computed as descriptors or fingerprints, or obtained as pretrained embeddings; TFMs can then serve as downstream predictors with extremely low computational cost and state-of-the-art performance~\cite{hollmannAccuratePredictionsSmall2025,quTabICLTabularFoundation2025,chenTabPFNOpensNew2026}.
This avoids the computational cost and complexity of fine-tuning molecular foundation models for every new task.

\begin{figure}
\centering
\includegraphics[width=1.\linewidth]{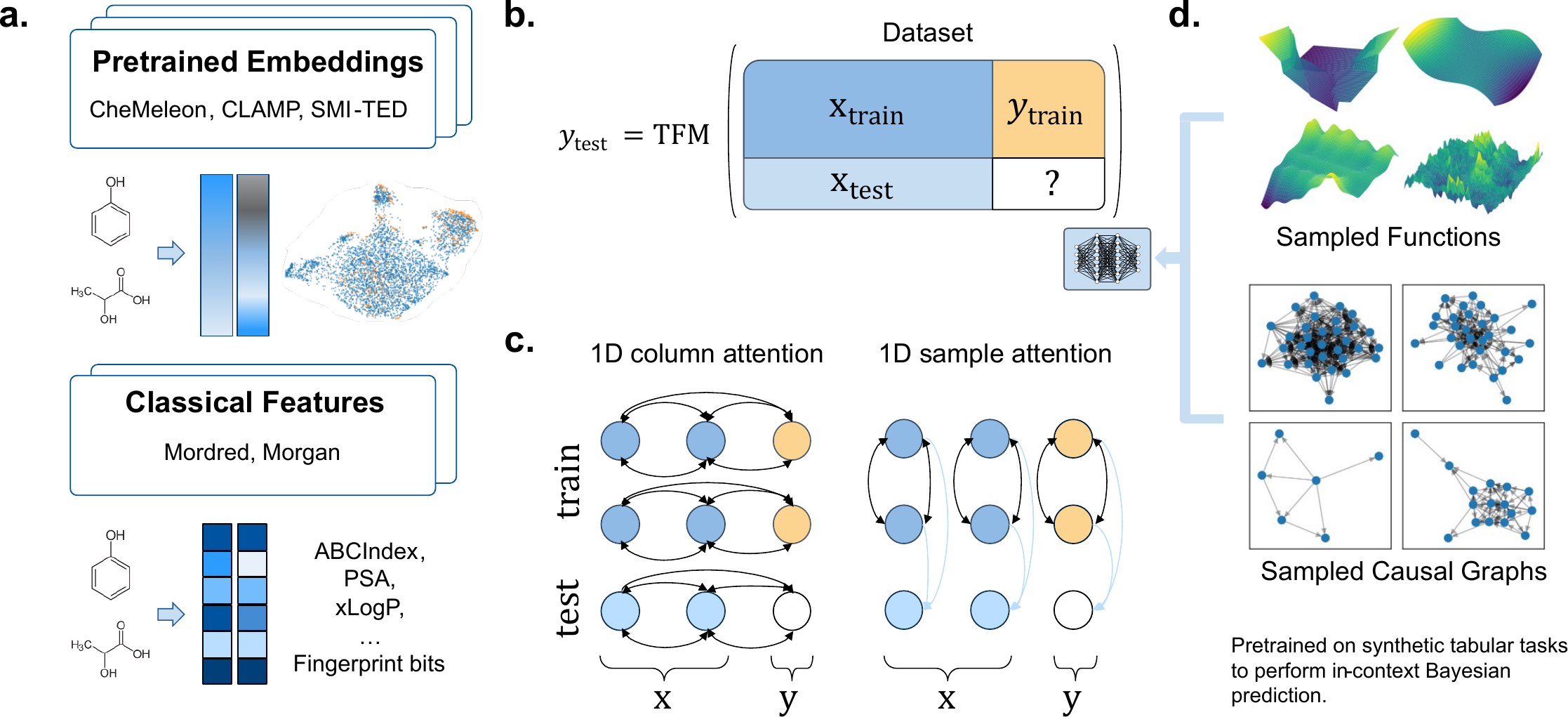}
\caption{Combining tabular foundation models (TFMs) with diverse molecular representations for property prediction: (a) Molecular representations used in this study, including pretrained embeddings from frozen foundation models (CheMeleon, CLAMP, and SMI-TED) and classical features (RDKit2d and Mordred descriptors, and Morgan fingerprints).
(b) In-context prediction with a tabular foundation model (TFM).
For each task, the training inputs and labels are provided together with the unlabeled test input at inference time, and the model predicts the missing target without task-specific training or fine-tuning.
(c) Schematic of the TFM architecture.
The model first applies one-dimensional attention over feature columns and then over samples, enabling joint processing of training and test examples.
The resulting representations are passed to a multilayer perceptron (MLP; not shown), which outputs either class probabilities for classification or predictive quantiles for regression.
(d) Synthetic-data pretraining paradigm for TFMs.
TFMs used in this work are pretrained exclusively on large collections of synthetic tabular tasks generated from sampled functions and structural causal models.
These expose the model to diverse statistical structures during pretraining, enabling amortized Bayesian inference on new datasets through in-context prediction alone.
Subplots in d adapted from~\cite{quTabICLv2BetterFaster2026}.}
\label{fig:fig1}
\end{figure}

We investigate the use of TFMs for molecular property prediction in the low- to medium-data regime by combining them with a diverse set of molecular representations, illustrated in Figure~\ref{fig:fig1}.
Specifically, we combine both TabPFN and TabICL with compact RDKit2d descriptors~\cite{rdkit2025}, larger Mordred descriptors~\cite{moriwakiMordredMolecularDescriptor2018}, Morgan fingerprints~\cite{rdkit2025}, and expressive pretrained embeddings from CheMeleon~\cite{burnsDeepLearningFoundation2026}, SMI-TED~\cite{soaresOpensourceFamilyLarge2025}, and CLAMP~\cite{seidlEnhancingActivityPrediction2023a}, thereby yielding 12 featurizer-model configurations, covering both mechanistic and foundation-model-derived feature spaces.

We benchmark against the best prior molecular baselines~\cite{burnsDeepLearningFoundation2026, heid2023chemprop, klaserMiniMolParameterEfficientFoundation2024, rossLargescaleChemicalLanguage2022} and strong classical ML models, including XGBoost~\cite{sheridanExtremeGradientBoosting2016} and CatBoost~\cite{prokhorenkova2018catboost} on 58 datasets from the Polaris and MoleculeACE benchmark suites~\cite{ashPracticallySignificantMethod2025,huangTherapeuticsDataCommons2021,burnsDeepLearningFoundation2026,vantilborgExposingLimitationsMolecular2022}, with a particular focus on the small- to medium-sized (up to $\sim$6000 samples) datasets that characterize many realistic applications, where data collection is expensive.
To further test TFMs in practical scenarios, we evaluate and compare them with highly tuned and domain specialized, state-of-the-art
literature baselines across 11 practical engineering datasets, including fuel ignition properties~\cite{schweidtmann2020graph, bin2025exploring}, polymer properties~\cite{kuenneth2023polybert}, and polymer-solvent interactions~\cite{liao2025directed}, thereby extending the analysis beyond predominantly pharmaceutical and bioactivity-focused public benchmarks.

This is the first demonstration that frozen molecular foundation model embeddings combined with tabular foundation models can outperform both classical machine learning and advanced fine-tuned deep learning foundation models across diverse molecular property prediction benchmarks.
TabPFN combined with expressive representations consistently outperforms prior state-of-the-art methods, including fine-tuned molecular foundation models such as CheMeleon, with up to 100\% win rates, defined as best or statistically indistinguishable from the best.
In contrast, prior work~\cite{pinto2025superior}, which used an older version of TabPFN primarily as a probe to identify pretrained foundation model layers suitable for fine-tuning, found that although its performance was generally correlated with downstream results across most tasks, it remained substantially weaker than fine-tuning in absolute terms.

We further show that, in contrast to recent work evaluating TabPFN for drug discovery~\cite{chenTabPFNOpensNew2026}, which indicated that ``TabPFN is largely invariant to representation choice'', we find that the choice of molecular representation is a major driver of performance, with CheMeleon embeddings and 2D descriptor sets performing most favorably and Morgan fingerprints performing substantially worse.

At the same time, TFMs require substantially less compute time in our runtime case study, with speedups of up to $27\times$ on CPU and $46\times$ on GPU, depending on dataset size and using 2048-dimensional CheMeleon embeddings (the largest feature set considered).
The same performance trend also extends to the engineering datasets, where TFM-based models remain competitive with highly hyperparameter-tuned literature baselines.

Our contribution demonstrates that in-context prediction with TFMs and suitable molecular representations provides a simple, accurate, and cost-efficient approach to real-world molecular property prediction.

\section{Results}
\subsection{Benchmarking on Polaris and MoleculeACE}

\noindent The Polaris~\cite{wognumCallIndustryledInitiative2024} and MoleculeACE~\cite{vantilborgExposingLimitationsMolecular2022} benchmark suites provide highly representative property prediction tasks in the low- to medium-data regime, comprising 58 tasks in total.
Polaris covers a broad range of practically relevant endpoints, including solubility, physiology, and biophysics, whereas MoleculeACE focuses on activity cliffs, a particularly challenging setting in which structurally similar molecules exhibit large differences in activity~\cite{vantilborgExposingLimitationsMolecular2022}.
We evaluate tabular foundation models (TFMs) with molecular descriptors and pretrained embeddings in these public benchmarks and compare them to various baselines, reporting win rates and average ranks across tasks. All tasks are evaluated on expert-curated fixed train-test splits across five random seeds; see Section~\ref{sec:Methods} for details.

\begin{table}[bhtp]
\centering
\caption{Aggregate performance across the Polaris and MoleculeACE benchmark suites (58 tasks total) for a subset of models evaluated on both benchmarks.}
\label{tab:polaris_moleculeace_winrates}
\begin{tabular}{lccc}
\toprule
Model & Win Count & Win Rate (\%) & Average Rank \\
\midrule
TabPFN-CheMeleonFP & 50 & 86.2 & 4.52 \\
TabPFN-RDKit2d & 33 & 56.9 & 5.31 \\
TabPFN-Mordred & 39 & 67.2 & 5.71 \\
TabICL-RDKit2d & 32 & 55.2 & 5.90 \\
TabICL-CheMeleonFP & 43 & 74.1 & 6.31 \\
TabICL-Mordred & 22 & 37.9 & 8.35 \\
CheMeleon & 24 & 41.4 & 8.84 \\
TabICL-CLAMP & 25 & 43.1 & 8.86 \\
TabPFN-CLAMP & 22 & 37.9 & 8.91 \\
minimol & 19 & 32.8 & 11.43 \\
CatBoost-Mordred & 11 & 19.0 & 12.05 \\
TabPFN-Morgan & 13 & 22.4 & 12.98 \\
RF Morgan & 10 & 17.2 & 15.12 \\
fastprop & 5 & 8.6 & 15.98 \\
MLP PLR Pretrained & 6 & 10.3 & 16.60 \\
RF Mordred & 6 & 10.3 & 17.50 \\
TabICL-Morgan & 8 & 13.8 & 17.64 \\
MoLFormer & 6 & 10.3 & 18.36 \\
PCA MLP Prefitted & 5 & 8.6 & 18.74 \\
XGBoost-Mordred & 6 & 10.3 & 19.53 \\
Chemprop & 2 & 3.4 & 21.74 \\
TabICL-SMI-TED & 4 & 6.9 & 24.09 \\
TabPFN-SMI-TED & 3 & 5.2 & 24.45 \\
MolCLR & 1 & 1.7 & 25.81 \\
\bottomrule
\end{tabular}
\footnotesize
Wins denote tasks on which a model is best or statistically indistinguishable from the best by Tukey’s HSD test over five random seeds; Win Rate gives the corresponding percentage of tasks. Average Rank is obtained by averaging within-dataset model ranks across all tasks, with lower values indicating stronger performance. All evaluations use the original expert-curated train-test splits and benchmark-specific metrics. See Section~\ref{sec:Methods} for evaluation and statistical details.
\end{table}

Across the combined benchmark suite, shown in Table~\ref{tab:polaris_moleculeace_winrates}, TFM-based approaches achieve stronger aggregate performance than the compared classical baselines and previously reported molecular foundation models.
The best overall configuration, TabPFN-CheMeleonFP (TabPFN with frozen CheMeleon embeddings), achieves 50 wins out of 58 tasks, corresponding to an 86.2\% win rate and an average rank of 4.52.
Among descriptor-based methods, TabPFN-RDKit2d achieves the next-best average rank, with a 56.9\% win rate and average rank of 5.31, while TabPFN-Mordred achieves a higher 67.2\% win rate and an average rank of 5.71.
TabICL-CheMeleonFP also remains strong, with a 74.1\% win rate and average rank of 6.31.
By comparison, the fine-tuned CheMeleon model reaches a win rate of 41.4\% and an average rank of 8.84 under the matched split and evaluation protocol~\cite{burnsDeepLearningFoundation2026}.
Thus, pairing TFMs with frozen molecular representations not only matches but substantially exceeds the performance of task-specific fine-tuning on these benchmarks.

\begin{figure}[htbp]
\centering
\includegraphics[width=0.95\linewidth]{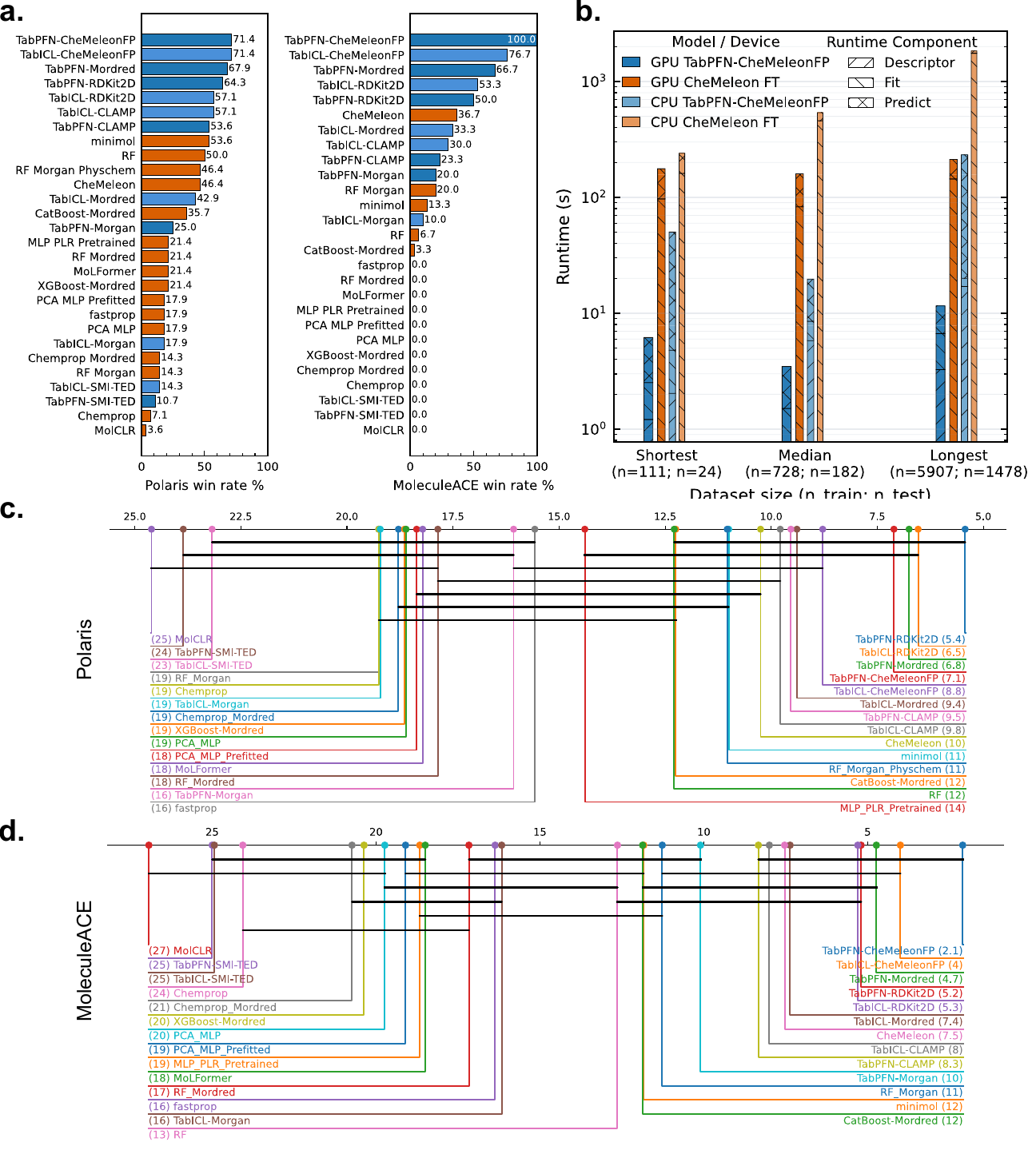}
\caption{
Performance ranking, win rates, and runtime across molecular property prediction benchmarks.
(a) Win rates for Polaris and MoleculeACE, defined as the percentage of tasks on which a model is best or statistically indistinguishable from the best using Tukey’s honestly significant difference test.
On Polaris, TabPFN-CheMeleonFP and TabICL-CheMeleonFP each reach 71.4\%, followed by TabPFN-Mordred (67.9\%) and TabPFN-RDKit2d (64.3\%), versus 46.4\% for CheMeleon; on MoleculeACE, TabPFN-CheMeleonFP reaches 100.0\%, followed by TabICL-CheMeleonFP (76.7\%), TabPFN-Mordred (66.7\%), TabICL-RDKit2d (53.3\%), TabPFN-RDKit2d (50.0\%), and CheMeleon (36.7\%).
TabPFN models are shown in dark blue, TabICL models in light blue, and previously reported baselines from the benchmark-matched comparison of~\cite{burnsDeepLearningFoundation2026} are shown in orange.
(b) Runtime decomposition for three representative Polaris datasets used as an illustrative case study.
Total runtime is separated into feature computation, fitting or training, and prediction, and compared between TabPFN-CheMeleonFP and fine-tuned CheMeleon on both GPU and CPU, with speedups of $4.8\times$--$27.3\times$ on CPU and $18.3\times$--$46.0\times$ on GPU.
(c) Critical-difference diagram using the Nemenyi post hoc test following Friedman’s test at a significance level of $\alpha=0.05$ for Polaris (28 tasks).
Lower rank indicates better mean performance.
Horizontal bars span models whose ranks are not significantly different in the post hoc comparison.
(d) Critical-difference diagram for MoleculeACE (30 tasks), with the same interpretation as in c.
}
\label{fig:polaris_moleace}
\end{figure}

The benchmark-wise analyses in Fig.~\ref{fig:polaris_moleace} show that this aggregate result is driven by gains on both suites.
On Polaris, the highest win rates are achieved by TabPFN-CheMeleonFP and TabICL-CheMeleonFP, each with 20 wins out of 28 tasks (71.4\%) under the win-rate definition above, followed by TabPFN-Mordred with 19 wins (67.9\%) and TabPFN-RDKit2d with 18 wins (64.3\%).
By average rank, the leading methods are TabPFN-RDKit2d (5.43), TabICL-RDKit2d (6.54), TabPFN-Mordred (6.75), and TabPFN-CheMeleonFP (7.11), compared with 10.25 for CheMeleon and 11.00 for minimol~\cite{burnsDeepLearningFoundation2026}.
On MoleculeACE, the separation is even clearer: TabPFN-CheMeleonFP is best or statistically tied for best on all 30 tasks (100.0\%) and achieves an average rank of 2.10, followed by TabICL-CheMeleonFP at a 76.7\% win rate and average rank of 4.00, and TabPFN-Mordred at 66.7\% and 4.73.
RDKit2d variants are also competitive on MoleculeACE, with TabICL-RDKit2d reaching a 53.3\% win rate and average rank of 5.30 and TabPFN-RDKit2d reaching 50.0\% and 5.20.
CheMeleon reaches a 36.7\% win rate and average rank of 7.53 on MoleculeACE, despite having been reported as highly competitive relative to earlier baselines~\cite{burnsDeepLearningFoundation2026}.
Notably, twelve methods record zero wins on MoleculeACE, highlighting the difficulty of this benchmark.
Despite TFMs strong general performance compared to CheMeleon, they still do not perform equally well for cliff and non-cliff molecules, with details given in Supplementary Fig.~\ref{fig:moleculeace_consistency}.
Benchmark-specific aggregate tables are provided in Supplementary Tables~\ref{tab:polaris_winrates} and~\ref{tab:moleculeace_winrates}, with the corresponding Tukey HSD summaries in Supplementary Figs.~\ref{fig:hsd_polaris} and~\ref{fig:hsd_moleculeace}.

Furthermore, consistent trends emerge in relation to TFM model choice and molecular representation.
TabPFN often outperforms TabICL for the same representation, most clearly for CheMeleonFP, RDKit2d, and Mordred, in agreement with recent results on general tabular benchmarks~\cite{ericksonTabArenaLivingBenchmark2025}.
Nevertheless, TabICL remains highly competitive: when paired with CheMeleon embeddings, it still outperforms or matches most previously reported baselines and remains close to the best-performing methods overall.
The molecular representation is another major determinant of downstream performance.
The strongest results are obtained with CheMeleon-derived embeddings and 2D descriptor sets, including both RDKit2d and Mordred, whereas Morgan fingerprint variants perform clearly worse.
For example, TabPFN-Morgan reaches only a 22.4\% win rate in the combined benchmark and has substantially lower aggregate performance than TabPFN-CheMeleonFP, TabPFN-RDKit2d, and TabPFN-Mordred.
Thus, the gains of TFMs are not representation-agnostic but depend strongly on sufficiently expressive molecular features.
Restricted feature-set comparisons, including RDKit2d-MACCS variants, are reported separately for Polaris and MoleculeACE in Supplementary Tables~\ref{tab:polaris_extra_features} and~\ref{tab:moleculeace_extra_features}.

It is particularly noteworthy that the configurations combining TFMs with CheMeleon-derived embeddings outperform the fine-tuned version of the CheMeleon model itself.
This emphasizes the predictive capabilities of TFMs when paired with expressive representations, further suggesting that task-specific fine-tuning of large foundation models is not always necessary to obtain strong results on small- to medium-sized downstream datasets.

The runtime analysis in Fig.~\ref{fig:polaris_moleace}b further shows that these performance gains are accompanied by substantially lower runtime cost.
Across representative Polaris datasets spanning the smallest (\texttt{adme-fang-rppb-1}, $n_{\mathrm{train}}=111$, $n_{\mathrm{test}}=24$), median (\texttt{caco2-wang}, $n_{\mathrm{train}}=728$, $n_{\mathrm{test}}=182$), and largest (\texttt{ld50-zhu}, $n_{\mathrm{train}}=5907$, $n_{\mathrm{test}}=1478$) training-set sizes, TabPFN-CheMeleonFP is consistently faster than fine-tuned CheMeleon on both GPU and CPU, yielding speedups of $4.8\times$--$27.3\times$ on CPU and $18.3\times$--$46.0\times$ on GPU. This is consistent with previously reported computational efficiency of TFMs~\cite{grinsztajn2025tabpfn,quTabICLv2BetterFaster2026}.

Overall, the benchmark results establish TFMs, especially TabPFN paired with strong molecular representations, as a simple and effective alternative to classical molecular property prediction methods and recent fine-tuning approaches based on molecular foundation models.

\subsection{Chemical Engineering Datasets}
Real-world chemical engineering datasets differ in important ways from the benchmarks considered above.
In particular, Polaris and MoleculeACE, while fairly broad and highly valuable as controlled test beds, are centered largely on pharmaceutical and bioactivity-related endpoints.
To test whether the strong benchmark performance of TFMs transfers beyond these applications, we therefore evaluate the same approach on three representative engineering-oriented applications: fuel ignition-property prediction, single-property polymer prediction, and polymer-solvent interaction prediction.
These datasets certainly also fall in the low-data regime similar to the previous benchmarks. Moreover, they introduce practical challenges that are common in engineering problems, including diverse physicochemical targets, inputs involving more than one molecule (i.e., molecular mixtures), and additional numerical input features, such as temperature.
As above, we combine out-of-the-box TFMs with fixed molecular representations and compare them with highly tuned and specialized, state-of-the-art literature baselines under the evaluation protocols used in the respective studies, with results summarized in Table~\ref{tab:engineering_combined}; detailed fuel, polymer, and polymer-solvent tables are provided in Supplementary Tables~\ref{tab:fpi_fuels_rmse}, \ref{tab:fpi_polymers_r2}, and~\ref{tab:fpi_interaction_r2}.
Reported literature performances are referenced with their respective sources in the tables.
Because several of the used literature baselines were not released with reliable fold-wise raw metrics, we interpret these experiments as contextual transfer tests rather than as a fully controlled benchmark.
Aggregate performance on the engineering tasks and per-task Tukey HSD plots are provided in Supplementary Table~\ref{tab:winrates_engineering} and Supplementary Fig.~\ref{fig:hsd_engineering}.

\begin{table*}[htbp]
\centering
\caption{Model performance on the chemical engineering datasets.}
\label{tab:engineering_combined}
\small
\setlength{\tabcolsep}{4pt}
\renewcommand{\arraystretch}{1.1}
\resizebox{\textwidth}{!}{%
\begin{tabular}{lccccccccccc}
\toprule
& \multicolumn{3}{c}{Fuels} & \multicolumn{7}{c}{Polymers} & \multicolumn{1}{c}{PolySolv} \\
\cmidrule(lr){2-4} \cmidrule(lr){5-11} \cmidrule(lr){12-12}
Metric & RMSE $\downarrow$ & RMSE $\downarrow$ & RMSE $\downarrow$ & $R^2$ $\uparrow$ & $R^2$ $\uparrow$ & $R^2$ $\uparrow$ & $R^2$ $\uparrow$ & $R^2$ $\uparrow$ & $R^2$ $\uparrow$ & $R^2$ $\uparrow$ & $R^2$ $\uparrow$ \\
Model & DCN & RON & MON & Eea & Egb & Egc & Ei & EPS & Nc & Xc & $\chi$ \\
\midrule
TabPFN-Mordred & \textbf{5.97 $\pm$ 1.69} & \textbf{8.57 $\pm$ 1.78} & 7.35 $\pm$ 1.62 & \textbf{0.93 $\pm$ 0.01} & \textbf{0.93 $\pm$ 0.01} & \textbf{0.93 $\pm$ 0.00} & 0.81 $\pm$ 0.02 & 0.78 $\pm$ 0.08 & 0.85 $\pm$ 0.06 & 0.39 $\pm$ 0.16 & 0.88 $\pm$ 0.05 \\
TabPFN-RDKit2d & 7.39 $\pm$ 1.75 & 9.43 $\pm$ 1.99 & 7.88 $\pm$ 2.42 & 0.92 $\pm$ 0.01 & \textbf{0.93 $\pm$ 0.01} & 0.92 $\pm$ 0.01 & 0.80 $\pm$ 0.03 & 0.77 $\pm$ 0.07 & 0.85 $\pm$ 0.06 & 0.40 $\pm$ 0.13 & \textbf{0.93 $\pm$ 0.03} \\
TabICL-RDKit2d & 6.81 $\pm$ 1.68 & 9.25 $\pm$ 1.82 & 7.67 $\pm$ 2.59 & \textbf{0.93 $\pm$ 0.01} & 0.92 $\pm$ 0.02 & 0.92 $\pm$ 0.01 & 0.81 $\pm$ 0.03 & 0.75 $\pm$ 0.06 & 0.84 $\pm$ 0.04 & 0.41 $\pm$ 0.14 & \textbf{0.93 $\pm$ 0.03} \\
TabICL-Mordred & 6.59 $\pm$ 1.30 & 9.02 $\pm$ 2.26 & \textbf{7.20 $\pm$ 1.69} & \textbf{0.93 $\pm$ 0.01} & 0.92 $\pm$ 0.02 & 0.91 $\pm$ 0.01 & 0.81 $\pm$ 0.03 & 0.77 $\pm$ 0.06 & 0.84 $\pm$ 0.03 & 0.39 $\pm$ 0.12 & 0.82 $\pm$ 0.06 \\
TabPFN-CheMeleonFP & 8.22 $\pm$ 1.97 & 9.68 $\pm$ 2.16 & 8.01 $\pm$ 2.22 & 0.91 $\pm$ 0.01 & 0.91 $\pm$ 0.02 & 0.92 $\pm$ 0.01 & 0.79 $\pm$ 0.01 & 0.77 $\pm$ 0.07 & 0.84 $\pm$ 0.05 & 0.41 $\pm$ 0.11 & 0.81 $\pm$ 0.08 \\
CatBoost-Mordred & 9.06 $\pm$ 1.76 & 9.12 $\pm$ 1.70 & 8.12 $\pm$ 1.97 & 0.90 $\pm$ 0.02 & 0.91 $\pm$ 0.01 & 0.91 $\pm$ 0.01 & 0.80 $\pm$ 0.03 & 0.75 $\pm$ 0.08 & 0.83 $\pm$ 0.05 & 0.38 $\pm$ 0.16 & 0.90 $\pm$ 0.03 \\
CheMeleon & 7.35 $\pm$ 1.95 & 9.48 $\pm$ 1.80 & 7.94 $\pm$ 2.36 & 0.91 $\pm$ 0.01 & 0.92 $\pm$ 0.01 & 0.91 $\pm$ 0.01 & 0.76 $\pm$ 0.01 & 0.72 $\pm$ 0.07 & 0.82 $\pm$ 0.04 & 0.22 $\pm$ 0.18 & 0.83 $\pm$ 0.08 \\
TabICL-CheMeleonFP & 8.60 $\pm$ 1.43 & 9.89 $\pm$ 1.91 & 8.28 $\pm$ 2.00 & 0.91 $\pm$ 0.01 & 0.91 $\pm$ 0.02 & 0.90 $\pm$ 0.01 & 0.78 $\pm$ 0.03 & 0.75 $\pm$ 0.05 & 0.82 $\pm$ 0.04 & 0.40 $\pm$ 0.11 & 0.71 $\pm$ 0.10 \\
CatBoost-RDKit2d & 9.56 $\pm$ 2.01 & 10.03 $\pm$ 1.82 & 8.39 $\pm$ 2.09 & 0.87 $\pm$ 0.02 & 0.89 $\pm$ 0.03 & 0.90 $\pm$ 0.01 & 0.78 $\pm$ 0.02 & 0.72 $\pm$ 0.08 & 0.80 $\pm$ 0.06 & 0.36 $\pm$ 0.18 & 0.90 $\pm$ 0.03 \\
CatBoost-CheMeleonFP & 11.11 $\pm$ 2.32 & 10.13 $\pm$ 2.26 & 9.31 $\pm$ 2.43 & 0.87 $\pm$ 0.02 & 0.88 $\pm$ 0.03 & 0.89 $\pm$ 0.01 & 0.77 $\pm$ 0.03 & 0.73 $\pm$ 0.05 & 0.81 $\pm$ 0.03 & 0.41 $\pm$ 0.09 & 0.88 $\pm$ 0.04 \\
TabPFN-Morgan & 11.24 $\pm$ 2.47 & 11.81 $\pm$ 2.27 & 10.25 $\pm$ 2.72 & 0.90 $\pm$ 0.01 & 0.88 $\pm$ 0.02 & 0.87 $\pm$ 0.01 & 0.78 $\pm$ 0.03 & 0.76 $\pm$ 0.05 & 0.83 $\pm$ 0.04 & 0.38 $\pm$ 0.12 & 0.88 $\pm$ 0.04 \\
TabICL-Morgan & 11.77 $\pm$ 2.40 & 11.99 $\pm$ 2.85 & 10.02 $\pm$ 3.11 & 0.90 $\pm$ 0.02 & 0.89 $\pm$ 0.01 & 0.84 $\pm$ 0.02 & 0.76 $\pm$ 0.03 & 0.77 $\pm$ 0.04 & 0.83 $\pm$ 0.02 & 0.37 $\pm$ 0.14 & 0.87 $\pm$ 0.04 \\
RF-Mordred & 11.83 $\pm$ 1.34 & 10.44 $\pm$ 1.64 & 9.26 $\pm$ 2.14 & 0.86 $\pm$ 0.03 & 0.88 $\pm$ 0.03 & 0.89 $\pm$ 0.01 & 0.76 $\pm$ 0.05 & 0.71 $\pm$ 0.08 & 0.79 $\pm$ 0.04 & 0.37 $\pm$ 0.10 & 0.82 $\pm$ 0.05 \\
XGBoost-Mordred & 10.68 $\pm$ 2.16 & 10.77 $\pm$ 1.35 & 9.86 $\pm$ 2.04 & 0.84 $\pm$ 0.04 & 0.88 $\pm$ 0.03 & 0.89 $\pm$ 0.01 & 0.76 $\pm$ 0.06 & 0.66 $\pm$ 0.09 & 0.77 $\pm$ 0.07 & 0.25 $\pm$ 0.20 & 0.87 $\pm$ 0.04 \\
RF-RDKit2d & 12.22 $\pm$ 2.03 & 10.84 $\pm$ 1.84 & 9.88 $\pm$ 2.19 & 0.82 $\pm$ 0.02 & 0.89 $\pm$ 0.02 & 0.88 $\pm$ 0.01 & 0.76 $\pm$ 0.05 & 0.69 $\pm$ 0.08 & 0.77 $\pm$ 0.06 & 0.38 $\pm$ 0.13 & 0.85 $\pm$ 0.03 \\
CatBoost-Morgan & 11.73 $\pm$ 2.55 & 12.01 $\pm$ 2.09 & 10.91 $\pm$ 2.50 & 0.88 $\pm$ 0.02 & 0.87 $\pm$ 0.03 & 0.87 $\pm$ 0.01 & 0.77 $\pm$ 0.03 & 0.75 $\pm$ 0.05 & 0.82 $\pm$ 0.05 & 0.35 $\pm$ 0.14 & 0.79 $\pm$ 0.06 \\
XGBoost-RDKit2d & 11.13 $\pm$ 1.71 & 11.54 $\pm$ 1.75 & 10.18 $\pm$ 2.08 & 0.83 $\pm$ 0.01 & 0.88 $\pm$ 0.03 & 0.88 $\pm$ 0.01 & 0.74 $\pm$ 0.06 & 0.67 $\pm$ 0.10 & 0.75 $\pm$ 0.08 & 0.30 $\pm$ 0.19 & 0.88 $\pm$ 0.04 \\
RF-CheMeleonFP & 12.50 $\pm$ 1.97 & 11.22 $\pm$ 2.42 & 10.47 $\pm$ 2.50 & 0.82 $\pm$ 0.04 & 0.86 $\pm$ 0.05 & 0.87 $\pm$ 0.01 & 0.75 $\pm$ 0.04 & 0.69 $\pm$ 0.08 & 0.78 $\pm$ 0.03 & 0.39 $\pm$ 0.06 & 0.74 $\pm$ 0.08 \\
RF-Morgan & 13.38 $\pm$ 2.10 & 13.03 $\pm$ 1.91 & 11.86 $\pm$ 2.92 & 0.83 $\pm$ 0.02 & 0.85 $\pm$ 0.05 & 0.87 $\pm$ 0.01 & 0.76 $\pm$ 0.03 & 0.74 $\pm$ 0.04 & 0.79 $\pm$ 0.04 & 0.33 $\pm$ 0.16 & 0.79 $\pm$ 0.04 \\
XGBoost-Morgan & 12.30 $\pm$ 2.45 & 12.87 $\pm$ 1.88 & 11.89 $\pm$ 1.93 & 0.85 $\pm$ 0.02 & 0.87 $\pm$ 0.04 & 0.87 $\pm$ 0.01 & 0.73 $\pm$ 0.03 & 0.72 $\pm$ 0.05 & 0.78 $\pm$ 0.05 & 0.29 $\pm$ 0.18 & 0.77 $\pm$ 0.07 \\
XGBoost-CheMeleonFP & 12.83 $\pm$ 1.92 & 12.33 $\pm$ 3.03 & 10.75 $\pm$ 2.75 & 0.80 $\pm$ 0.06 & 0.87 $\pm$ 0.04 & 0.88 $\pm$ 0.01 & 0.75 $\pm$ 0.02 & 0.66 $\pm$ 0.07 & 0.77 $\pm$ 0.03 & 0.29 $\pm$ 0.10 & 0.80 $\pm$ 0.06 \\
Chemprop GNN & 13.39 $\pm$ 2.12 & 11.89 $\pm$ 2.10 & 12.50 $\pm$ 3.13 & 0.82 $\pm$ 0.04 & 0.86 $\pm$ 0.02 & 0.90 $\pm$ 0.01 & 0.67 $\pm$ 0.05 & 0.65 $\pm$ 0.10 & 0.72 $\pm$ 0.06 & 0.22 $\pm$ 0.06 & 0.72 $\pm$ 0.09 \\
\midrule
\multicolumn{12}{l}{\textit{Literature baselines}} \\
MTL-Train~\cite{bin2025exploring} & 7.52 $\pm$ 1.84 & 9.46 $\pm$ 1.72 & 8.47 $\pm$ 1.95 & -- & -- & -- & -- & -- & -- & -- & -- \\
STL~\cite{bin2025exploring} & 7.51 $\pm$ 1.86 & 10.29 $\pm$ 2.16 & 8.83 $\pm$ 2.19 & -- & -- & -- & -- & -- & -- & -- & -- \\
PolyCL~\cite{zhou2025polycl} & -- & -- & -- & 0.91 & 0.89 & 0.88 & \textbf{0.81} & \textbf{0.79} & \textbf{0.85} & 0.40 & -- \\
TransPolymer~\cite{xu2023transpolymer} & -- & -- & -- & 0.89 & 0.90 & 0.88 & 0.79 & 0.76 & 0.81 & \textbf{0.46} & -- \\
PolyBERT~\cite{kuenneth2023polybert} & -- & -- & -- & 0.91 & 0.88 & 0.88 & 0.77 & 0.77 & 0.80 & 0.44 & -- \\
D-MPNN-TC~\cite{liao2025directed} & -- & -- & -- & -- & -- & -- & -- & -- & -- & -- & \textbf{0.93 $\pm$ 0.03} \\
\bottomrule
\end{tabular}%
}
\footnotesize
Values are given as mean $\pm$ standard deviation where available, aggregated over ten folds for the fuels and PolySolv datasets and over five folds for the polymer dataset. Fuel tasks are evaluated by RMSE, where lower values are better; polymer and PolySolv tasks are evaluated by $R^2$, where higher values are better. Bold entries indicate the best displayed mean performance in each task; if displayed means are tied, the displayed standard deviation is used as a tie-breaker, and entries with identical displayed mean and standard deviation are all bold. Literature values without fold-wise standard deviations are treated as having the lowest standard deviation for this tie-breaking only.
\end{table*}

The fuel ignition benchmarks evaluate how well models can predict experimentally measured combustion-quality indicators such as DCN, RON, and MON, which are central to fuel design because they directly reflect ignition and anti-knock performance in engines~\cite{schweidtmann2020graph}.
On these tasks, TabPFN-Mordred achieves the lowest mean RMSE across ten folds on the DCN and RON benchmarks.
TabICL-Mordred achieves the lowest mean RMSE on the MON task. 
The GNN MTL-Train~\cite{bin2025exploring} exploits correlations between fuel targets, whereas our methods are evaluated as single-task models, and it performs slightly worse on average than CheMeleon.

The polymer-property benchmarks cover a diverse set of electronic (Eea, Egb, Egc, Ei), dielectric (EPS, Nc), and structural (Xc) targets, which are frequently required in the design of novel polymers~\cite{zhou2025polycl,xu2023transpolymer,kuenneth2023polybert}.
This makes them important test cases for assessing whether models can generalize across practically relevant material properties in low-data settings.
Polymers are featurized via their repeat unit SMILES.
Here, TFM-based models, especially TabPFN-Mordred and the RDKit2d variants, match or exceed the strongest reported baselines on several targets, yielding the best mean performance, or tying the best reported literature value, on Eea, Egb, Egc, Ei, and Nc while remaining competitive on EPS. Xc is the only target on which the TFM-based models do not match the best reported literature performance, although they still remain competitive with it.

The polymer–solvent interaction benchmark (PolySolv) focuses on predicting Flory–Huggins $\chi$ interaction parameters for polymer–solvent pairs, a challenging and practically important task because miscibility strongly influences formulation, processing, and polymer product design.
In addition to the repeat-unit SMILES and the solvent SMILES, the models also receive temperature and volume fraction as additional input features.
On the PolySolv benchmark, the highly tuned and specialized D-MPNN-TC baseline of Liao et al.~\cite{liao2025directed} achieves the top reported $R^2$, while TabPFN-RDKit2d matches this value within rounding and the untuned CatBoost-Mordred baseline and several other TFM variants achieve accuracy close to it.

\begin{figure}
\centering
\includegraphics[width=0.85\linewidth]{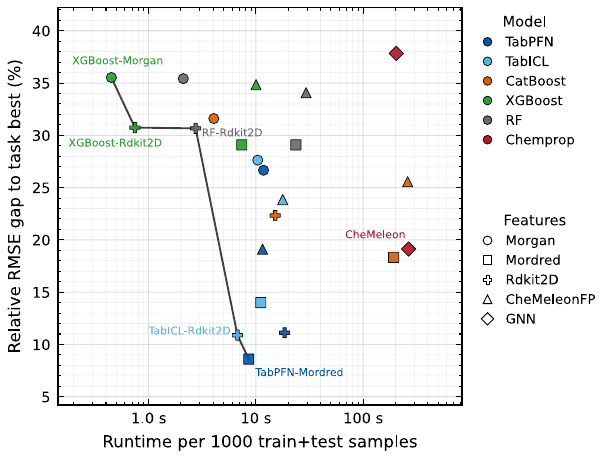}
\caption{Average runtime per 1,000 samples across the 11 engineering tasks and folds, measured as training or fitting plus inference for 1,000 samples and excluding feature computation, plotted against relative RMSE gap. Relative RMSE gap is defined as the RMSE reduction relative to the best method on that task.}
\label{fig:pareto_runtime}
\end{figure}

Figure~\ref{fig:pareto_runtime} shows a Pareto plot of the relative RMSE gap to the best model on a task, defined as $\frac{\text{RMSE}-\text{RMSE}_{\text{best}}}{\text{RMSE}}$ versus the average runtime per 1,000 total train and test samples, averaged across the 11 engineering tasks.
Both runtime and relative RMSE gap were computed fold-wise under the respective cross-validation protocol and then averaged across folds. 
The x-axis reports runtime per 1,000 samples, defined as model fitting or training plus prediction, excluding feature computation. 
Feature computation is especially costly for Mordred descriptors, which places the frozen foundation model embeddings and the smaller RDKit2d descriptor set at a practical advantage.
The resulting plot reveals a clear Pareto front in the trade-off between computational cost and relative predictive performance. 
In particular, XGBoost occupies the extreme low-cost end, delivering very fast predictions, whereas TabPFN combined with RDKit2d descriptors and, especially, Mordred descriptors achieves the strongest overall performance. 
By contrast, the fine-tuned CheMeleon baseline has an average relative RMSE gap that is about 10 percentage points larger while also being an order of magnitude slower.

Overall, our results suggest that the strong performance of TFMs is not limited to the controlled Polaris and MoleculeACE benchmarks with a predominantly biological or pharmaceutical focus, but can also carry over to engineering-relevant molecular prediction tasks.
Across polymer, polymer-solvent, and fuel datasets, the TFM-based pipeline remains competitive with substantially more specialized baselines, supporting its use as a simple and effective approach in realistic engineering applications.

\section{Discussion}
We investigate a new paradigm for molecular property prediction: TFMs paired with frozen molecular foundation model embeddings and 2D descriptor sets, thereby enabling in-context learning.
Across two representative benchmark suites including datasets from a few hundred to several thousand property data points, the TFM-based approach clearly outperforms the evaluated classical prediction methods and fine-tuned molecular foundation models, in terms of aggregate predictive performance, while being substantially faster.
The strong performance also transfers to practical chemical engineering datasets, where TFMs are competitive with extensively hyperparameter-tuned, domain-specialized models from the literature.

The practical significance of our results is straightforward: because small- to medium-sized datasets are not only the most common but also extremely practically/industrially relevant setting in molecular property prediction, TFMs combined with in-context learning and rich molecular representations provide an important advance by improving both predictive performance and efficiency while also being easy to deploy.
In many real-world projects, the bottleneck is not only predictive performance, but the time, compute, and specialized expertise needed to build and maintain task-specific deep learning pipelines.
The presented approach reduces this burden to a simple two-step procedure: compute a high-quality molecular representation once, then use a pretrained tabular foundation model as the predictor.
This yields a cheaper, faster, and more accessible workflow for applications such as drug discovery, catalysis, and process design.

Our findings also have broader implications for how molecular foundation models should be used and evaluated.
Much recent work has focused on end-to-end downstream fine-tuning as the default way to extract value from pretrained models.
Our results suggest that, at least in the small- to medium-data regime, the most effective way to leverage molecular foundation model representations may not be through gradient-based fine-tuning, which prior work has shown can overfit~\cite{tetkoBeAwareOverfitting2024}, but rather through pairing frozen representations with in-context learning and TFMs.
This shifts the emphasis from task-specific adaptation toward representation quality, predictor robustness, and workflow efficiency.
In that sense, TFMs provide a practical mechanism for turning molecular foundation model embeddings into strong predictors without requiring the additional cost and fragility of per-task optimization.

At the same time, our study has substantial room for improvement.
First, our benchmarks focus primarily on small- to medium-sized supervised datasets, and it remains to be established how well the observed advantages persist for substantially larger datasets, where end-to-end training becomes more competitive.
Although we also explored larger-scale datasets, these experiments became hardware-constrained for larger feature sets, making them difficult to benchmark, even though the smaller RDKit2d features continued to run without issues.
Specifically, they became memory-limited without disk offloading and disk-limited with disk offloading ($\sim 50\,\mathrm{GB}$ per fold), such that the practical benefits of TFM-based pipelines over standard deep learning fine-tuning, especially simplicity, ease of use, and speed, were substantially reduced.
As detailed in Supplementary Information Section~\ref{apdx:brouwer_activity_coefficients}, the Brouwer activity-coefficient dataset comprises approximately 21,000 measurements, and performance remains strong for compact pair representations such as RDKit2d, whereas higher-dimensional representations start to approach practical limits.
Second, the present work considers primarily single-molecule prediction tasks; extensions to chemically more complex settings, such as many-component mixtures, remain open and potentially challenging, since TFM computation scales considerably with feature dimension~\cite{hollmannAccuratePredictionsSmall2025, quTabICLv2BetterFaster2026}.
Third, we have focused on out-of-the-box use of TFMs without changing any inference hyperparameters.
A simple way to boost performance is to increase the number of estimators for TabPFN, because TabPFN subsamples features for each estimator; see Supplementary Information Section~\ref{apdx:model_details} and Supplementary Table~\ref{tab:tfm_prediction_hyperparameters}.

These limitations define several promising directions for future research.
It will be important to test the approach systematically on larger datasets, multimolecular systems, and tasks where structured inductive biases or physical constraints matter~\cite{rittig2025molecular}.
In addition, we have only considered a single set of frozen representations per model; hybrid strategies combining in-context prediction with ensembling~\cite{erickson2020autogluon} and fine-tuning where necessary could yield further gains, mostly at the cost of extra compute.
Another interesting extension is multi-task prediction.
TabPFN and TabICL are designed and pretrained for single-task prediction and can only indirectly exploit cross-task dependencies without again resorting to gradient-based fine-tuning~\cite{sinodinos2026multitask}.
Native support for in-context multitask learning~\cite{sinodinos2026multitab} in future TFMs would therefore be highly valuable, because many applications require to predict multiple targets at the same time with shared underlying physics or biology.
Prior work has shown that uncertainty estimates from TFMs can be surprisingly well calibrated out-of-the-box~\cite{hollmannTabPFNTransformerThat2023}.
This opens the door to combining TFMs with uncertainty-aware decision-making strategies, such as active learning and Bayesian optimization~\cite{yuGITBOHighDimensionalBayesian2026}, in applications like solvent design, where experimental data are limited and data-efficient exploration is especially valuable.
More broadly, our results motivate the development of molecular ML pipelines that optimize not only for raw accuracy but also for computational cost, simplicity, and reliability under realistic deployment constraints.

We expect TFMs to be valuable both as a practical route to deployment in real-world applications and as a strong baseline for assessing the predictive performance of pretrained foundation models in chemistry.

\section{Methods}\label{sec:Methods}
\subsection{Tabular In-Context Learning}
TFMs are transformer-based models that, in the versions used here, are pretrained exclusively on large collections of synthetic tabular tasks generated from randomly sampled functions and structural causal models (SCMs)~\cite{mullerTransformersCanBayesian2024,hollmannTabPFNTransformerThat2023,quTabICLv2BetterFaster2026}.
During pretraining, tasks are sampled continuously rather than reused, exposing the model to a broad range of statistical dependencies and causal structures.
Formally, if a prediction task $\mathcal{T}=(\mathcal{D}_{\mathrm{train}},\mathcal{D}_{\mathrm{test}})$ is sampled from an SCM-induced prior $p(\mathcal{T})$, the model parameters are optimized according to
\begin{align}
\mathcal{L}(\theta) = \mathbb{E}_{\mathcal{T} \sim p(\mathcal{T})} \left[
- \log q_\theta(y_{\mathrm{test}} \mid x_{\mathrm{test}}, \mathcal{D}_{\mathrm{train}})
\right].
\end{align}

This objective trains the model to predict the label of a query example conditional on a set of in-context training examples, rather than through task-specific parameter updates.
At inference, the labeled training inputs and outputs for a benchmark task are presented together with an unlabeled test input, and the TFM predicts the missing target directly in context, without retraining or fine-tuning.

TFMs process each input table using one-dimensional sequential attention over feature columns and then over samples.
The resulting latent representations are passed to a multilayer perceptron head, which outputs class probabilities for classification tasks and predictive quantiles or binned densities for regression tasks.
This synthetic pretraining paradigm can be interpreted as amortizing or learning Bayesian inference over a broad prior family of tabular problems, enabling the model to adapt to new datasets from examples alone~\cite{mullerTransformersCanBayesian2024}.
For architectural and theoretical details, we refer the reader to~\cite{garg2025real,hollmannAccuratePredictionsSmall2025,quTabICLv2BetterFaster2026}.

Recent iterations have substantially increased the scale and diversity of pretraining tasks, leading to strong empirical performance on public tabular benchmarks such as TabArena~\cite{ericksonTabArenaLivingBenchmark2025,quTabICLv2BetterFaster2026}.

In the present study, the TFMs are used strictly in in-context inference mode with fixed molecular feature vectors; the TFMs are never retrained or fine-tuned on any molecular benchmark and are not exposed to molecule-specific data during pretraining.

\subsection{Molecular Representations}
The classical molecular representations used in this study are RDKit2d descriptors~\cite{rdkit2025}, 2D Mordred descriptors~\cite{moriwakiMordredMolecularDescriptor2018}, and Morgan fingerprints (radius 2, 2048 bits).
In supplementary feature comparisons, we additionally evaluate RDKit2d descriptors concatenated with MACCS keys.
The frozen pretrained embeddings are computed with CheMeleon, SMI-TED~\cite{soaresOpensourceFamilyLarge2025}, and CLAMP~\cite{seidlEnhancingActivityPrediction2023a}.
CheMeleon is a descriptor-pretrained molecular foundation model based on a directed message-passing neural network and trained to predict deterministic Mordred descriptors, thereby learning transferable molecular embeddings for downstream property prediction~\cite{burnsDeepLearningFoundation2026}.
Throughout the manuscript, \texttt{CheMeleonFP} denotes the fixed embedding returned by the released CheMeleon message-passing network before task-specific fine-tuning.
SMI-TED is a large-scale SMILES encoder-decoder foundation model trained with self-supervision on PubChem molecules to produce general-purpose molecular embeddings~\cite{soaresOpensourceFamilyLarge2025}.
CLAMP is a contrastive multimodal model that aligns molecular structure with assay text descriptions and yields transferable molecular embeddings for low-data assay activity prediction~\cite{seidlEnhancingActivityPrediction2023a}.
The corresponding feature dimensions of each model are listed in Supplementary Table~\ref{tab:features}.

\subsection{Models}
We evaluate TabPFN-2.6~\cite{priorlabs_tabpfn_github,grinsztajn2025tabpfn,hollmannAccuratePredictionsSmall2025} and TabICL~\cite{quTabICLv2BetterFaster2026} with the above-mentioned classical and frozen foundation model representations.
We compare these results with previously reported classical baselines and fine-tuned molecular foundation models, including CheMeleon~\cite{burnsDeepLearningFoundation2026}, minimol~\cite{klaserMiniMolParameterEfficientFoundation2024}, MolFormer~\cite{rossLargescaleChemicalLanguage2022}, and others evaluated by~\cite{burnsDeepLearningFoundation2026}, using the released predictions from their GitHub repository.
In addition, we extend the benchmarks to include untuned XGBoost and CatBoost with Mordred descriptors~\cite{sheridanExtremeGradientBoosting2016,prokhorenkova2018catboost}.
In the engineering datasets, we also combine the classical tabular ML baselines with foundation model embeddings.

\subsection{Preprocessing}
All featurization and scaling steps are fit on the training portion of each dataset only and then applied unchanged to the test set.
Descriptor features and frozen embeddings are all standardized using the training-set feature means and standard deviations and then clamped to the range $[-6,6]$ to limit the effect of extreme values, following~\cite{burnsDeepLearningFoundation2026}.
The considered TFMs apply further normalization or outlier removal steps, which we leave in place to assess out-of-the-box performance without changes to hyperparameters.
RDKit2d descriptors are computed from the RDKit descriptor list.
Mordred descriptors are computed with the 2D Mordred descriptor set (\texttt{ignore\_3D=True}); Morgan fingerprints are computed directly from the parsed RDKit molecule.
CheMeleonFP features are generated from the released frozen CheMeleon model, with Chemprop version 2.2.1~\cite{burnsDeepLearningFoundation2026, heid2023chemprop}.
For regression tasks, targets are standardized using the training-set mean and variance and predictions are transformed back to the original scale before benchmark evaluation.
For Polaris classification tasks, predicted probabilities are used directly for scoring, with hard labels obtained by thresholding at 0.5 when required by the benchmark interface~\cite{burnsDeepLearningFoundation2026}.

\subsection{Benchmarks}
To enable a direct and fair comparison with prior work, we evaluate our approach across the same benchmark suites used in previous studies, namely Polaris and MoleculeACE~\cite{burnsDeepLearningFoundation2026, ashPracticallySignificantMethod2025, huangTherapeuticsDataCommons2021,vantilborgExposingLimitationsMolecular2022}.
These comprise 28 and 30 datasets, respectively, and are representative of practical molecular property prediction tasks.
The Polaris suite covers curated endpoints spanning solubility, ADMET, biophysics, and physiology from Polaris and Therapeutics Data Commons, and each dataset is evaluated with its benchmark-specific primary metric~\cite{ashPracticallySignificantMethod2025,huangTherapeuticsDataCommons2021}.
MoleculeACE consists of 30 ChEMBL activity prediction assays with predefined expert-guided train-test splits based on ECFP similarity and activity, thereby explicitly probing generalization across activity cliffs~\cite{vantilborgExposingLimitationsMolecular2022}.

\subsubsection{Performance Evaluation}
Following~\cite{burnsDeepLearningFoundation2026}, we consider only single-task prediction problems and strictly reuse the original train-test splits.
Each newly run benchmark experiment is repeated over five random seeds to quantify stochastic variation.
We reuse the train-test splits and evaluation setup from~\cite{burnsDeepLearningFoundation2026}, and therefore report their main CheMeleon results together with the other published baseline results to enable a split- and metric-matched comparison.
We do not perform task-specific hyperparameter optimization for any of the classical tabular ML models, because the goal is to assess practical out-of-the-box performance.
The TFMs used are TabPFN-2.6~\cite{priorlabs_tabpfn_github, grinsztajn2025tabpfn,hollmannAccuratePredictionsSmall2025} and TabICLv2~\cite{quTabICLv2BetterFaster2026}.
For MoleculeACE, we compute the overall test RMSE as well as cliff and non-cliff subset RMSEs from the same predictions; the overall test RMSE is used for the aggregated model ranking shown in Fig.~\ref{fig:polaris_moleace}.
For Polaris, benchmark scores are taken directly from the Polaris evaluation interface using the dataset-specific main metric.
When the Polaris metric is an error quantity, such as mean squared error or mean absolute error, we rescale scores within each dataset to the unit interval and invert them so that larger values always indicate better performance before rank aggregation~\cite{burnsDeepLearningFoundation2026}.

We use Tukey's honest significant difference (HSD) test with $\alpha=0.05$ on the five seed-specific benchmark scores to determine whether a model is statistically indistinguishable from the best model on a given dataset, following prior work~\cite{ashPracticallySignificantMethod2025,burnsDeepLearningFoundation2026}.
Because these repeated seeds share the same fixed train-test split, this test quantifies variation due to model stochasticity rather than uncertainty over alternative molecular test sets.

The top model and all models not significantly different from it are counted as wins, and win counts are aggregated into the reported win rates.
In parallel, we compute within-dataset model ranks and average them across datasets.
The critical-difference diagrams in Fig.~\ref{fig:polaris_moleace}c,d are based on these average ranks and use the Nemenyi post hoc comparison following Friedman's test at $\alpha=0.05$.

\subsubsection{Runtime Case Study}
To quantify computational cost, we perform an illustrative runtime case study comparing TabPFN-CheMeleonFP with fine-tuned CheMeleon on three Polaris datasets selected to represent the smallest, median, and largest training-set sizes: \texttt{adme-fang-rppb-1} ($n_{\mathrm{train}}=111$), \texttt{caco2-wang} ($n_{\mathrm{train}}=728$), and \texttt{ld50-zhu} ($n_{\mathrm{train}}=5907$).
For the TabPFN-CheMeleonFP pipeline, total runtime is decomposed into CheMeleon embedding generation on the training and test molecules, TFM fitting, and test-set prediction.
For the fine-tuned CheMeleon baseline, total runtime is decomposed into Chemprop fine-tuning from the CheMeleon foundation weights and prediction on the test set.
Both CPU and GPU runtimes are measured with the same execution path and a fixed random seed; CPU measurements are obtained on an Intel Xeon 8468 Sapphire system using 24 cores, and GPU measurements are obtained on a single NVIDIA H100.

\subsection{Chemical Engineering Datasets}
\subsubsection{Data}
To assess transfer beyond the Polaris and MoleculeACE benchmark suites, we additionally evaluate the methods on 11 chemical engineering regression datasets spanning fuel-property prediction, polymer-property prediction, and polymer-solvent interaction tasks.

\textbf{Fuels.} The fuel ignition dataset comprises 505 molecules annotated with up to three experimentally measured fuel-performance properties: derived cetane number (DCN), research octane number (RON), and motor octane number (MON)~\cite{schweidtmann2020graph, bin2025exploring}.
These properties probe different aspects of ignition and knocking behavior, and the dataset is incomplete because not every molecule has measurements for all three targets.

\textbf{Polymers.} We consider seven polymer-property datasets covering chain band gap (Egc), bulk band gap (Egb), dielectric constant (EPS), refractive index (Nc), ionization energy (Ei), electron affinity (Eea), and crystallization tendency (Xc)~\cite{kuenneth2023polybert}.
Together, these datasets span optoelectronic, dielectric, and structural polymer properties.
All targets except Xc are obtained from density functional theory calculations, whereas Xc is derived from experimental heats of fusion combined with a group-contribution method~\cite{kuenneth2023polybert}.

\textbf{PolySolv.} In addition, we consider the PolySolv dataset, which contains 1,208 experimental Flory-Huggins interaction parameters ($\chi$) for 38 polymers and 150 solvents~\cite{mark2007physical, liao2025directed}.
This dataset captures experimentally measured and regressed polymer–solvent interactions across a chemically diverse set of combinations.

\subsubsection{Accuracy Evaluation and Literature Baselines}
All considered TFMs are evaluated with RDKit2d, Morgan, Mordred, and CheMeleonFP representations, alongside graph-based Chemprop baselines, fine-tuned CheMeleon, and untuned random forest, XGBoost, and CatBoost baselines on the same feature spaces.
Here, Chemprop GNN denotes a standard Chemprop message-passing network, and CheMeleon denotes fine-tuning the CheMeleon foundation model through the Chemprop training pipeline.
To remain comparable with the respective literature baselines, we repeat the benchmark-specific cross-validation protocols.

\textbf{Fuels.} 
For the fuel tasks, we report the mean and standard deviation of RMSE across folds.
We compare against the literature baselines GNN STL and GNN MTL-Train from~\cite{bin2025exploring}, with matched random-split 10-fold cross-validation.

\textbf{Polymers.}
Polymers are passed to the respective featurizers using repeat-unit SMILES notation.
For the polymer-property benchmarks, we report the mean and standard deviation of $R^2$ across folds.
For comparison, we use the published PolyCL, TransPolymer, and PolyBERT results reported in the corresponding studies~\cite{zhou2025polycl,xu2023transpolymer,kuenneth2023polybert}, with matched random-split 5-fold cross-validation.
These do not include fold-wise standard deviations.

\textbf{PolySolv.}
For the PolySolv benchmark, polymer and solvent molecules are featurized separately, and their feature vectors are concatenated with the numerical covariates temperature and volume fraction before regression.
For Chemprop GNN and CheMeleon, the same two SMILES columns and numerical covariates are passed to the model.
We report the mean and standard deviation of $R^2$ across folds.
We compare against the published D-MPNN-TC baseline from~\cite{liao2025directed}.
\section*{Data Availability}
No new phyiscal experimental data were generated in this work. The datasets used are available from their respective authors via Polaris or GitHub. Processed benchmark inputs, predictions, and analysis outputs, as well as predictions from the CheMeleon repository~\cite{burnsDeepLearningFoundation2026}, are available in our Git repository at \url{https://git.rwth-aachen.de/avt-svt/public/tabpfn-molprop}. A snapshot of the repository is also available on Zenodo~\cite{ben_hicham_2026_19605631}.
\section*{Code Availability}
The full code for this work including all implementations, experiments and analysis scripts is available under the open-source Eclipse Public License at \url{https://git.rwth-aachen.de/avt-svt/public/tabpfn-molprop}.

\bibliography{literature}

\section*{Author Contributions}
Karim K. Ben Hicham: Conceptualization, Methodology, Formal analysis, Investigation, Writing -- original draft.
Jan G. Rittig: Conceptualization, Supervision, Writing -- original draft.
Martin Grohe: Methodology, Funding acquisition, Writing -- review \& editing.
Alexander Mitsos: Conceptualization, Supervision, Funding acquisition, Writing -- review \& editing.

\section*{Acknowledgements}
We acknowledge support of the Werner Siemens Foundation in the frame of the WSS Research Center
``catalaix''.
We also thank the authors of CheMeleon~\cite{burnsDeepLearningFoundation2026} for releasing the full code and predictions for their work.

\section*{Competing Interests}
The authors declare no competing interests.

\clearpage

\setcounter{section}{0}
\setcounter{subsection}{0}
\setcounter{table}{0}
\setcounter{figure}{0}
\renewcommand{\thesection}{S\arabic{section}}
\renewcommand{\thesubsection}{\thesection.\arabic{subsection}}
\renewcommand{\thetable}{S\arabic{table}}
\renewcommand{\thefigure}{S\arabic{figure}}
\renewcommand{\theHsection}{S\arabic{section}}
\renewcommand{\theHsubsection}{S\arabic{section}.\arabic{subsection}}
\renewcommand{\theHtable}{S\arabic{table}}
\renewcommand{\theHfigure}{S\arabic{figure}}

\section{Featurization and Model Details}\label{apdx:model_details}
\textbf{TFM Inference Hyperparameter Defaults.} Tabular foundation models (TFMs) expose a range of inference-time hyperparameters. For TabICL, these include in particular the number of estimators, preprocessing and normalization choices, feature shuffling, and outlier-handling options. For TabPFN, relevant hyperparameters likewise include the number of estimators, normalization settings, handling of categorical features, softmax temperature, and outlier-removal steps.

In this work, however, we deliberately do not tune any of these hyperparameters and instead use both TFMs strictly with their default settings throughout. This choice is motivated by the central aim of this study: to evaluate TFMs as simple, out-of-the-box downstream predictors for molecular property prediction. We therefore focus on the user experience and practical performance that can be achieved without task-specific optimization, manual intervention, or additional machine-learning expertise.

This choice should yield conservative performance estimates. In particular, further improvements are likely possible through simple changes in the hyperparameters. For example, in the high-dimensional feature regimes considered here, increasing the number of estimators is recommended, especially for TabPFN, where individual estimators operate on subsets of 680 features for categorical features and 500 for numeric features, and predictions are aggregated across estimators. Such tuning could therefore improve feature coverage and downstream predictive performance for very high-dimensional molecular representations. Selected values are shown in Table~\ref{tab:tfm_prediction_hyperparameters}; details can be found in the respective works and GitHub repositories.

\textbf{Classical Features.} RDKit2d descriptors~\cite{rdkit2025}, Mordred descriptors~\cite{moriwakiMordredMolecularDescriptor2018}, and Morgan fingerprints~\cite{rdkit2025} are three common molecular representations.
RDKit2d descriptors provide a compact set of physicochemical and topological features and are widely used as a simple, robust baseline.
Mordred descriptors extend this to a much larger and more diverse descriptor set, offering broader structural coverage at the cost of higher dimensionality.
Morgan fingerprints, by contrast, encode local chemical substructures as binary or count-based patterns rather than continuous descriptors.
The RDKit2d-MACCS variant used in supplementary analyses augments RDKit2d descriptors with MACCS structural keys.
In practice, RDKit2d is the most compact descriptor baseline, Mordred is the most expressive descriptor representation, and Morgan fingerprints provide a sparse substructure-based alternative.

\begin{table}[htbp]
\centering
\small
\setlength{\tabcolsep}{6pt}
\renewcommand{\arraystretch}{1.05}
\caption{Default TFM inference hyperparameters.}
\label{tab:tfm_prediction_hyperparameters}
\begin{tabular}{lcc}
\toprule
Hyperparameter & TabICLv2 & TabPFN-2.6 \\
\midrule
\texttt{n\_estimators} & 8 & 8 \\
\texttt{outlier\_threshold} & 4.0 & \texttt{None} \\
\texttt{softmax\_temperature} & 0.9 & 0.9 \\
\texttt{prediction\_threshold} & 0.5 & 0.5 \\
\texttt{max\_features\_per\_estimator} & \texttt{all\_unique} & 680/500 \\
\texttt{feat\_shuffle\_method} & \texttt{latin} & \texttt{shuffle} \\
\bottomrule
\end{tabular}
\end{table}

\begin{table}[htbp]
\centering
\caption{Molecular representations used in the Polaris and MoleculeACE benchmark experiments.}
\label{tab:features}
\begin{tabular}{lr}
\toprule
Featurizer & Dimension\\
\midrule
CheMeleonFP~\cite{burnsDeepLearningFoundation2026} & 2048\\
CLAMP~\cite{seidlEnhancingActivityPrediction2023a} & 768 \\
SMI-TED~\cite{soaresOpensourceFamilyLarge2025} & 768 \\
RDKit2d~\cite{rdkit2025} & 217 \\
RDKit2d-MACCS~\cite{rdkit2025} & 383 \\
Mordred~\cite{moriwakiMordredMolecularDescriptor2018} & 1613 \\
Morgan (radius 2)~\cite{rdkit2025} & 2048 \\
\bottomrule
\end{tabular}
\end{table}

\section{Supplementary Benchmark Results}

\begin{table}[htbp]
\centering
\caption{Win rates and average ranks across the Polaris benchmark tasks. Wins count models that are best or statistically indistinguishable from the best model on a task.}
\begin{tabular}{lccc}
\toprule
Model & Win Count & Win Rate (\%) & Average Rank \\
\midrule
TabPFN-RDKit2d & 18 & 64.3 & 5.43 \\
TabICL-RDKit2d & 16 & 57.1 & 6.54 \\
TabPFN-Mordred & 19 & 67.9 & 6.75 \\
TabPFN-CheMeleonFP & 20 & 71.4 & 7.11 \\
TabICL-CheMeleonFP & 20 & 71.4 & 8.79 \\
TabICL-Mordred & 12 & 42.9 & 9.39 \\
TabPFN-CLAMP & 15 & 53.6 & 9.54 \\
TabICL-CLAMP & 16 & 57.1 & 9.79 \\
CheMeleon~\cite{burnsDeepLearningFoundation2026} & 13 & 46.4 & 10.25 \\
minimol~\cite{burnsDeepLearningFoundation2026} & 15 & 53.6 & 11.00 \\
RF Morgan Physchem~\cite{burnsDeepLearningFoundation2026} & 13 & 46.4 & 11.04 \\
CatBoost-Mordred & 10 & 35.7 & 12.25 \\
RF~\cite{burnsDeepLearningFoundation2026} & 14 & 50.0 & 12.29 \\
MLP PLR Pretrained~\cite{burnsDeepLearningFoundation2026} & 6 & 21.4 & 14.39 \\
fastprop~\cite{burnsDeepLearningFoundation2026} & 5 & 17.9 & 15.57 \\
TabPFN-Morgan & 7 & 25.0 & 16.07 \\
RF Mordred~\cite{burnsDeepLearningFoundation2026} & 6 & 21.4 & 17.86 \\
MoLFormer~\cite{burnsDeepLearningFoundation2026} & 6 & 21.4 & 18.21 \\
PCA MLP Prefitted~\cite{burnsDeepLearningFoundation2026} & 5 & 17.9 & 18.36 \\
PCA MLP~\cite{burnsDeepLearningFoundation2026} & 5 & 17.9 & 18.61 \\
XGBoost-Mordred & 6 & 21.4 & 18.64 \\
Chemprop Mordred~\cite{burnsDeepLearningFoundation2026} & 4 & 14.3 & 18.79 \\
TabICL-Morgan & 5 & 17.9 & 19.21 \\
RF Morgan~\cite{burnsDeepLearningFoundation2026} & 4 & 14.3 & 19.25 \\
Chemprop~\cite{burnsDeepLearningFoundation2026} & 2 & 7.1 & 19.25 \\
TabICL-SMI-TED & 4 & 14.3 & 23.18 \\
TabPFN-SMI-TED & 3 & 10.7 & 23.86 \\
MolCLR~\cite{burnsDeepLearningFoundation2026} & 1 & 3.6 & 24.61 \\
\bottomrule
\end{tabular}
\label{tab:polaris_winrates}
\end{table}

\begin{table}[htbp]
\centering
\caption{Win rates and average ranks across the MoleculeACE benchmark tasks. Wins count models that are best or statistically indistinguishable from the best model on a task.}
\begin{tabular}{lccc}
\toprule
Model & Win Count & Win Rate (\%) & Average Rank \\
\midrule
TabPFN-CheMeleonFP & 30 & 100.0 & 2.10 \\
TabICL-CheMeleonFP & 23 & 76.7 & 4.00 \\
TabPFN-Mordred & 20 & 66.7 & 4.73 \\
TabPFN-RDKit2d & 15 & 50.0 & 5.20 \\
TabICL-RDKit2d & 16 & 53.3 & 5.30 \\
TabICL-Mordred & 10 & 33.3 & 7.37 \\
CheMeleon~\cite{burnsDeepLearningFoundation2026} & 11 & 36.7 & 7.53 \\
TabICL-CLAMP & 9 & 30.0 & 8.00 \\
TabPFN-CLAMP & 7 & 23.3 & 8.33 \\
TabPFN-Morgan & 6 & 20.0 & 10.10 \\
RF Morgan~\cite{burnsDeepLearningFoundation2026} & 6 & 20.0 & 11.27 \\
minimol~\cite{burnsDeepLearningFoundation2026} & 4 & 13.3 & 11.83 \\
CatBoost-Mordred & 1 & 3.3 & 11.87 \\
RF~\cite{burnsDeepLearningFoundation2026} & 2 & 6.7 & 12.63 \\
TabICL-Morgan & 3 & 10.0 & 16.17 \\
fastprop~\cite{burnsDeepLearningFoundation2026} & 0 & 0.0 & 16.37 \\
RF Mordred~\cite{burnsDeepLearningFoundation2026} & 0 & 0.0 & 17.17 \\
MoLFormer~\cite{burnsDeepLearningFoundation2026} & 0 & 0.0 & 18.50 \\
MLP PLR Pretrained~\cite{burnsDeepLearningFoundation2026} & 0 & 0.0 & 18.67 \\
PCA MLP Prefitted~\cite{burnsDeepLearningFoundation2026} & 0 & 0.0 & 19.10 \\
PCA MLP~\cite{burnsDeepLearningFoundation2026} & 0 & 0.0 & 19.73 \\
XGBoost-Mordred & 0 & 0.0 & 20.37 \\
Chemprop Mordred~\cite{burnsDeepLearningFoundation2026} & 0 & 0.0 & 20.73 \\
Chemprop~\cite{burnsDeepLearningFoundation2026} & 0 & 0.0 & 24.07 \\
TabICL-SMI-TED & 0 & 0.0 & 24.93 \\
TabPFN-SMI-TED & 0 & 0.0 & 25.00 \\
MolCLR~\cite{burnsDeepLearningFoundation2026} & 0 & 0.0 & 26.93 \\
\bottomrule
\end{tabular}
\label{tab:moleculeace_winrates}
\end{table}

\begin{table}[htbp]
\centering
\caption{Restricted feature-set comparison across the MoleculeACE benchmark tasks, including RDKit2d-MACCS variants. Wins count models that are best or statistically indistinguishable from the best model within this comparison on a task.}
\begin{tabular}{lccc}
\toprule
Model & Win Count & Win Rate (\%) & Average Rank \\
\midrule
TabPFN-CheMeleonFP & 20 & 66.7 & 2.53 \\
TabICL-CheMeleonFP & 11 & 36.7 & 4.43 \\
TabPFN-Mordred & 6 & 20.0 & 5.37 \\
TabPFN-RDKit2d-MACCS & 13 & 43.3 & 5.60 \\
TabICL-RDKit2d-MACCS & 5 & 16.7 & 5.97 \\
TabPFN-RDKit2d & 12 & 40.0 & 6.07 \\
TabICL-RDKit2d & 5 & 16.7 & 6.07 \\
CheMeleon & 6 & 20.0 & 7.83 \\
TabICL-Mordred & 2 & 6.7 & 8.27 \\
TabICL-CLAMP & 2 & 6.7 & 8.77 \\
TabPFN-CLAMP & 4 & 13.3 & 8.87 \\
TabPFN-Morgan & 1 & 3.3 & 10.53 \\
RF & 0 & 0.0 & 11.97 \\
TabICL-Morgan & 1 & 3.3 & 13.03 \\
XGBoost-Mordred & 0 & 0.0 & 14.80 \\
TabICL-SMI-TED & 0 & 0.0 & 16.43 \\
TabPFN-SMI-TED & 0 & 0.0 & 16.47 \\
\bottomrule
\end{tabular}
\label{tab:moleculeace_extra_features}
\end{table}

\begin{table}[htbp]
\centering
\caption{Restricted feature-set comparison across the Polaris benchmark tasks, including RDKit2d-MACCS variants. Wins count models that are best or statistically indistinguishable from the best model within this comparison on a task.}
\begin{tabular}{lccc}
\toprule
Model & Win Count & Win Rate (\%) & Average Rank \\
\midrule
TabPFN-RDKit2d & 15 & 53.6 & 5.29 \\
TabPFN-RDKit2d-MACCS & 11 & 39.3 & 5.50 \\
TabPFN-Mordred & 11 & 39.3 & 5.86 \\
TabICL-RDKit2d & 9 & 32.1 & 6.14 \\
TabPFN-CheMeleonFP & 10 & 35.7 & 6.21 \\
TabICL-RDKit2d-MACCS & 11 & 39.3 & 6.75 \\
TabICL-CheMeleonFP & 8 & 28.6 & 7.54 \\
TabPFN-CLAMP & 8 & 28.6 & 7.61 \\
TabICL-CLAMP & 9 & 32.1 & 7.75 \\
CheMeleon~\cite{burnsDeepLearningFoundation2026} & 8 & 28.6 & 8.32 \\
TabICL-Mordred & 4 & 14.3 & 8.46 \\
RF~\cite{burnsDeepLearningFoundation2026} & 6 & 21.4 & 9.71 \\
TabPFN-Morgan & 3 & 10.7 & 11.32 \\
XGBoost-Mordred & 1 & 3.6 & 13.00 \\
TabICL-Morgan & 3 & 10.7 & 13.07 \\
TabICL-SMI-TED & 1 & 3.6 & 15.11 \\
TabPFN-SMI-TED & 1 & 3.6 & 15.36 \\
\bottomrule
\end{tabular}
\label{tab:polaris_extra_features}
\end{table}

\begin{figure}
\centering
\includegraphics[width=0.85\linewidth]{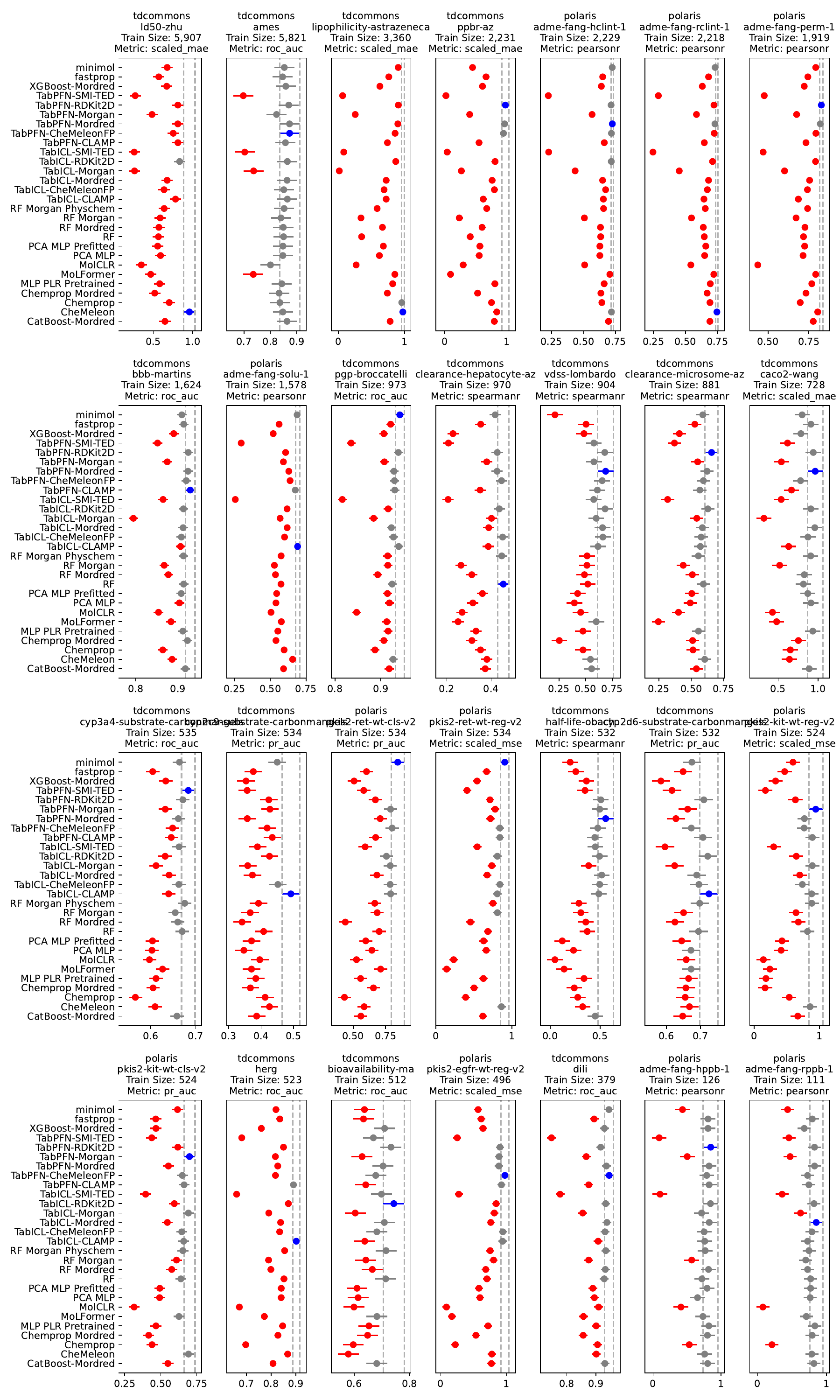}
\caption{Performance for all models for the 28 Polaris tasks. Models highlighted in blue are the top performers on a given benchmark, whereas models shown in gray are not significantly different from the top performer according to Tukey’s honestly significant difference (HSD) test ($\alpha = 0.05$) across five repetitions. Models shown in red performed significantly worse and were considered to have lost on that benchmark.}
\label{fig:hsd_polaris}
\end{figure}

\begin{figure}
\centering
\includegraphics[width=0.85\linewidth]{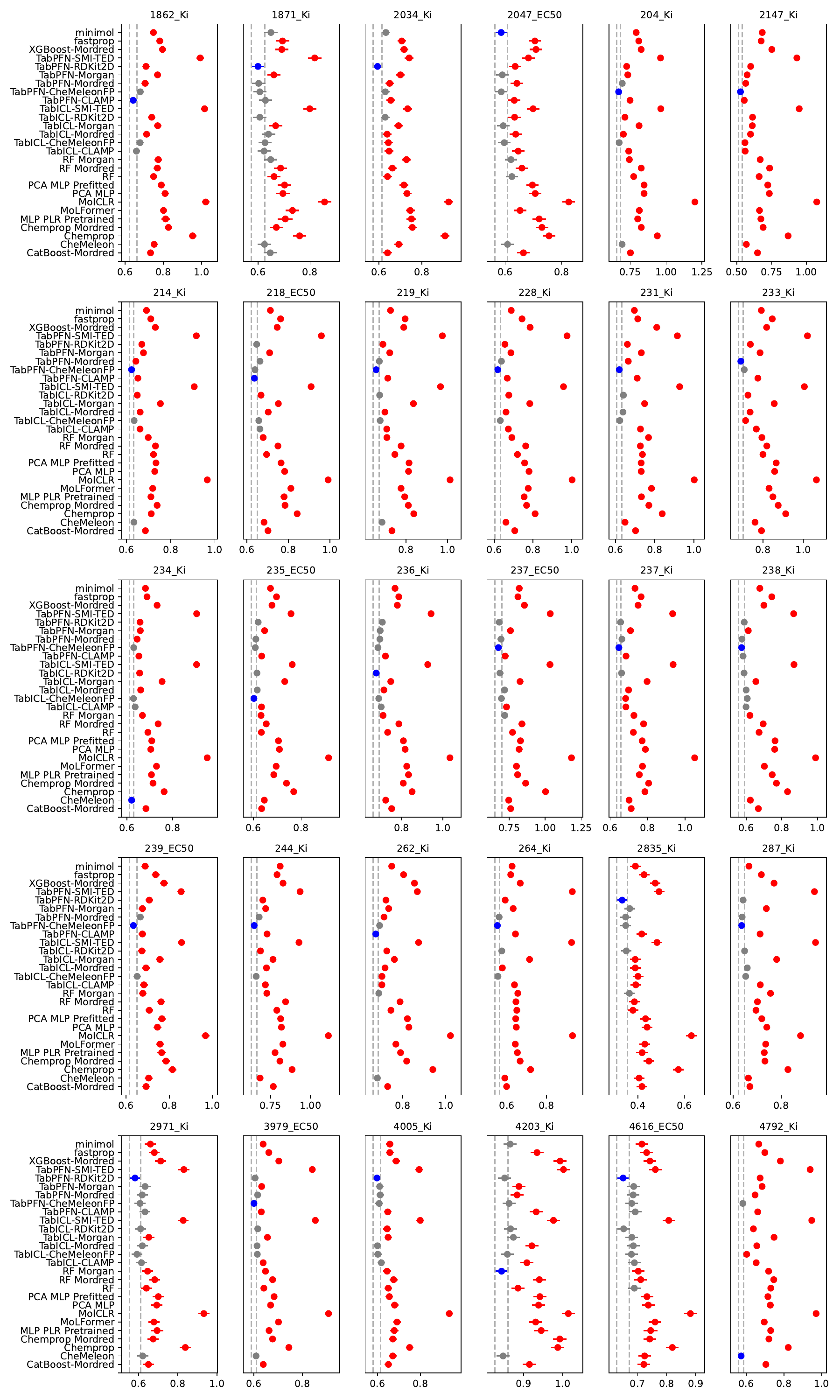}
\caption{Performance for all models for the 30 MoleculeACE tasks. Models highlighted in blue are the top performers on a given benchmark, whereas models shown in gray are not significantly different from the top performer according to Tukey’s honestly significant difference (HSD) test ($\alpha = 0.05$) across five repetitions. Models shown in red performed significantly worse and were considered to have lost on that benchmark.}
\label{fig:hsd_moleculeace}
\end{figure}

\begin{figure}
\centering
\includegraphics[width=0.85\linewidth]{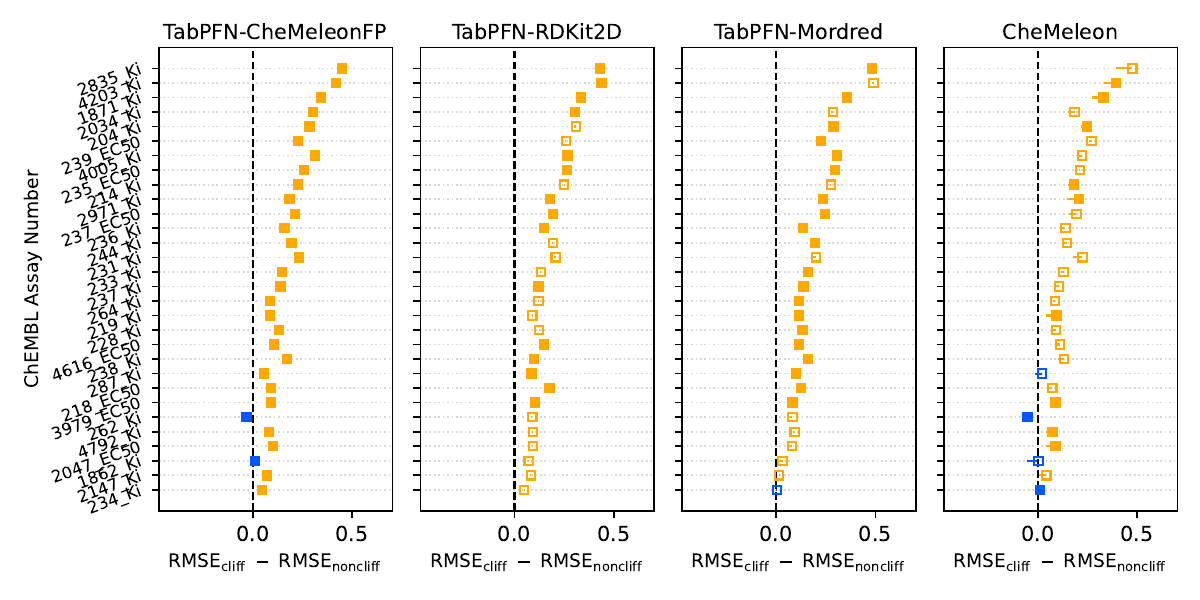}
\caption{Assay-level difference between prediction error on activity-cliff and non-cliff compounds across MoleculeACE benchmarks. Shown are values of $\mathrm{RMSE}_{\mathrm{cliff}} - \mathrm{RMSE}_{\mathrm{noncliff}}$ for TabPFN-CheMeleonFP, TabPFN-RDKit2d, TabPFN-Mordred, and CheMeleon across individual ChEMBL assays. Positive values indicate higher prediction error on activity-cliff compounds relative to non-cliff compounds, whereas values near zero indicate comparable performance across the two subsets. Filled markers indicate assays for which the corresponding model was the best-performing method or statistically indistinguishable from the best-performing method. Blue markers indicate assays for which $\mathrm{RMSE}_{\mathrm{cliff}} - \mathrm{RMSE}_{\mathrm{noncliff}}$ was not significantly different from zero.}
\label{fig:moleculeace_consistency}
\end{figure}

\textbf{Activity Cliffs.} Across MoleculeACE ChEMBL assays, the TabPFN-CheMeleonFP model showed particularly strong predictive performance overall, frequently matching or exceeding the best-performing methods at the assay level. Nevertheless, despite this strong overall performance, it still exhibited predominantly positive values of $\mathrm{RMSE}_{\mathrm{cliff}} - \mathrm{RMSE}_{\mathrm{noncliff}}$, indicating that activity-cliff compounds remained systematically more difficult to predict than non-cliff compounds, as shown in Figure~\ref{fig:moleculeace_consistency}.
TabPFN-CheMeleonFP displayed a limitation comparable to that observed for CheMeleon, with neither approach overcoming the prediction difficulty associated with activity cliffs. Accordingly, only 2 of 30 tasks for TabPFN-CheMeleonFP showed no significant difference between cliff and non-cliff RMSE, compared with 4 of 30 tasks for CheMeleon.
The magnitude of this cliff-associated error increase varied across assays and models. Filled markers denote assays for which a model was either the best-performing method or statistically indistinguishable from the best-performing method, whereas blue markers denote assays for which $\mathrm{RMSE}_{\mathrm{cliff}} - \mathrm{RMSE}_{\mathrm{noncliff}}$ was not significantly different from zero.

\newpage
\section{Supplementary Engineering Benchmark Results}
\begin{table*}[htbp]
\centering
\caption{Fuel benchmark 10-fold cross-validation RMSE results across datasets (mean $\pm$ standard deviation). Lower values indicate better performance.}
\label{tab:fpi_fuels_rmse}
\small
\setlength{\tabcolsep}{4pt}
\renewcommand{\arraystretch}{1.1}
\resizebox{0.8\textwidth}{!}{%
\begin{tabular}{lccc}
\toprule
Model & DCN & RON & MON \\
\midrule

TabPFN-Mordred & \textbf{5.97 $\pm$ 1.69} & \textbf{8.57 $\pm$ 1.78} & 7.35 $\pm$ 1.62 \\
TabICL-Mordred & 6.59 $\pm$ 1.30 & 9.02 $\pm$ 2.26 & \textbf{7.20 $\pm$ 1.69} \\
TabICL-RDKit2d & 6.81 $\pm$ 1.68 & 9.25 $\pm$ 1.82 & 7.67 $\pm$ 2.59 \\
TabPFN-RDKit2d & 7.39 $\pm$ 1.75 & 9.43 $\pm$ 1.99 & 7.88 $\pm$ 2.42 \\
CheMeleon & 7.35 $\pm$ 1.95 & 9.48 $\pm$ 1.80 & 7.94 $\pm$ 2.36 \\
CatBoost-Mordred & 9.06 $\pm$ 1.76 & 9.12 $\pm$ 1.70 & 8.12 $\pm$ 1.97 \\
TabPFN-CheMeleonFP & 8.22 $\pm$ 1.97 & 9.68 $\pm$ 2.16 & 8.01 $\pm$ 2.22 \\
GNN MTL-Train~\cite{bin2025exploring} & 7.52 $\pm$ 1.84 & 9.46 $\pm$ 1.72 & 8.47 $\pm$ 1.95 \\
TabICL-CheMeleonFP & 8.60 $\pm$ 1.43 & 9.89 $\pm$ 1.91 & 8.28 $\pm$ 2.00 \\
GNN STL~\cite{bin2025exploring} & 7.51 $\pm$ 1.86 & 10.29 $\pm$ 2.16 & 8.83 $\pm$ 2.19 \\
CatBoost-RDKit2d & 9.56 $\pm$ 2.01 & 10.03 $\pm$ 1.82 & 8.39 $\pm$ 2.09 \\
CatBoost-CheMeleonFP & 11.11 $\pm$ 2.32 & 10.13 $\pm$ 2.26 & 9.31 $\pm$ 2.43 \\
XGBoost-Mordred & 10.68 $\pm$ 2.16 & 10.77 $\pm$ 1.35 & 9.86 $\pm$ 2.04 \\
RF-Mordred & 11.83 $\pm$ 1.34 & 10.44 $\pm$ 1.64 & 9.26 $\pm$ 2.14 \\
XGBoost-RDKit2d & 11.13 $\pm$ 1.71 & 11.54 $\pm$ 1.75 & 10.18 $\pm$ 2.08 \\
RF-RDKit2d & 12.22 $\pm$ 2.03 & 10.84 $\pm$ 1.84 & 9.88 $\pm$ 2.19 \\
TabPFN-Morgan & 11.24 $\pm$ 2.47 & 11.81 $\pm$ 2.27 & 10.25 $\pm$ 2.72 \\
TabICL-Morgan & 11.77 $\pm$ 2.40 & 11.99 $\pm$ 2.85 & 10.02 $\pm$ 3.11 \\
RF-CheMeleonFP & 12.50 $\pm$ 1.97 & 11.22 $\pm$ 2.42 & 10.47 $\pm$ 2.50 \\
CatBoost-Morgan & 11.73 $\pm$ 2.55 & 12.01 $\pm$ 2.09 & 10.91 $\pm$ 2.50 \\
XGBoost-CheMeleonFP & 12.83 $\pm$ 1.92 & 12.33 $\pm$ 3.03 & 10.75 $\pm$ 2.75 \\
XGBoost-Morgan & 12.30 $\pm$ 2.45 & 12.87 $\pm$ 1.88 & 11.89 $\pm$ 1.93 \\
Chemprop GNN & 13.39 $\pm$ 2.12 & 11.89 $\pm$ 2.10 & 12.50 $\pm$ 3.13 \\
RF-Morgan & 13.38 $\pm$ 2.10 & 13.03 $\pm$ 1.91 & 11.86 $\pm$ 2.92 \\
\bottomrule
\end{tabular}%
}

\end{table*}

\begin{table*}[htbp]
\centering
\caption{Polymer benchmark 5-fold cross-validation $R^2$ results across datasets (mean $\pm$ standard deviation) with baseline rows. Boldface follows the rounded-mean and standard-deviation tie-breaking rule from Table~\ref{tab:engineering_combined}.}
\label{tab:fpi_polymers_r2}
\small
\setlength{\tabcolsep}{4pt}
\renewcommand{\arraystretch}{1.1}
\resizebox{\textwidth}{!}{%
\begin{tabular}{lllllllll}
\toprule
Model & Eea & Egb & Egc & Ei & EPS & Nc & Xc \\
\midrule
TabPFN-Mordred & \textbf{0.93 $\pm$ 0.01} & \textbf{0.93 $\pm$ 0.01} & \textbf{0.93 $\pm$ 0.00} & 0.81 $\pm$ 0.02 & 0.78 $\pm$ 0.08 & 0.85 $\pm$ 0.06 & 0.39 $\pm$ 0.16 \\
TabPFN-RDKit2d & 0.92 $\pm$ 0.01 & \textbf{0.93 $\pm$ 0.01} & 0.92 $\pm$ 0.01 & 0.80 $\pm$ 0.03 & 0.77 $\pm$ 0.07 & 0.85 $\pm$ 0.06 & 0.40 $\pm$ 0.13 \\
TabICL-RDKit2d & \textbf{0.93 $\pm$ 0.01} & 0.92 $\pm$ 0.02 & 0.92 $\pm$ 0.01 & 0.81 $\pm$ 0.03 & 0.75 $\pm$ 0.06 & 0.84 $\pm$ 0.04 & 0.41 $\pm$ 0.14 \\
TabICL-Mordred & \textbf{0.93 $\pm$ 0.01} & 0.92 $\pm$ 0.02 & 0.91 $\pm$ 0.01 & 0.81 $\pm$ 0.03 & 0.77 $\pm$ 0.06 & 0.84 $\pm$ 0.03 & 0.39 $\pm$ 0.12 \\
TabPFN-CheMeleonFP & 0.91 $\pm$ 0.01 & 0.91 $\pm$ 0.02 & 0.92 $\pm$ 0.01 & 0.79 $\pm$ 0.01 & 0.77 $\pm$ 0.07 & 0.84 $\pm$ 0.05 & 0.41 $\pm$ 0.11 \\
PolyCL~\cite{zhou2025polycl} & 0.91 & 0.89 & 0.88 & \textbf{0.81} & \textbf{0.79} & \textbf{0.85} & 0.40 \\
CatBoost-Mordred & 0.90 $\pm$ 0.02 & 0.91 $\pm$ 0.01 & 0.91 $\pm$ 0.01 & 0.80 $\pm$ 0.03 & 0.75 $\pm$ 0.08 & 0.83 $\pm$ 0.05 & 0.38 $\pm$ 0.16 \\
TabICL-CheMeleonFP & 0.91 $\pm$ 0.01 & 0.91 $\pm$ 0.02 & 0.90 $\pm$ 0.01 & 0.78 $\pm$ 0.03 & 0.75 $\pm$ 0.05 & 0.82 $\pm$ 0.04 & 0.40 $\pm$ 0.11 \\
TransPolymer~\cite{xu2023transpolymer} & 0.89 & 0.90 & 0.88 & 0.79 & 0.76 & 0.81 & \textbf{0.46} \\
PolyBERT~\cite{kuenneth2023polybert} & 0.91 & 0.88 & 0.88 & 0.77 & 0.77 & 0.80 & 0.44 \\
CheMeleon & 0.91 $\pm$ 0.01 & 0.92 $\pm$ 0.01 & 0.91 $\pm$ 0.01 & 0.76 $\pm$ 0.01 & 0.72 $\pm$ 0.07 & 0.82 $\pm$ 0.04 & 0.22 $\pm$ 0.18 \\
TabPFN-Morgan & 0.90 $\pm$ 0.01 & 0.88 $\pm$ 0.02 & 0.87 $\pm$ 0.01 & 0.78 $\pm$ 0.03 & 0.76 $\pm$ 0.05 & 0.83 $\pm$ 0.04 & 0.38 $\pm$ 0.12 \\
CatBoost-CheMeleonFP & 0.87 $\pm$ 0.02 & 0.88 $\pm$ 0.03 & 0.89 $\pm$ 0.01 & 0.77 $\pm$ 0.03 & 0.73 $\pm$ 0.05 & 0.81 $\pm$ 0.03 & 0.41 $\pm$ 0.09 \\
TabICL-Morgan & 0.90 $\pm$ 0.02 & 0.89 $\pm$ 0.01 & 0.84 $\pm$ 0.02 & 0.76 $\pm$ 0.03 & 0.77 $\pm$ 0.04 & 0.83 $\pm$ 0.02 & 0.37 $\pm$ 0.14 \\
CatBoost-RDKit2d & 0.87 $\pm$ 0.02 & 0.89 $\pm$ 0.03 & 0.90 $\pm$ 0.01 & 0.78 $\pm$ 0.02 & 0.72 $\pm$ 0.08 & 0.80 $\pm$ 0.06 & 0.36 $\pm$ 0.18 \\
RF-Mordred & 0.86 $\pm$ 0.03 & 0.88 $\pm$ 0.03 & 0.89 $\pm$ 0.01 & 0.76 $\pm$ 0.05 & 0.71 $\pm$ 0.08 & 0.79 $\pm$ 0.04 & 0.37 $\pm$ 0.10 \\
CatBoost-Morgan & 0.88 $\pm$ 0.02 & 0.87 $\pm$ 0.03 & 0.87 $\pm$ 0.01 & 0.77 $\pm$ 0.03 & 0.75 $\pm$ 0.05 & 0.82 $\pm$ 0.05 & 0.35 $\pm$ 0.14 \\
RF-RDKit2d & 0.82 $\pm$ 0.02 & 0.89 $\pm$ 0.02 & 0.88 $\pm$ 0.01 & 0.76 $\pm$ 0.05 & 0.69 $\pm$ 0.08 & 0.77 $\pm$ 0.06 & 0.38 $\pm$ 0.13 \\
RF-Morgan & 0.83 $\pm$ 0.02 & 0.85 $\pm$ 0.05 & 0.87 $\pm$ 0.01 & 0.76 $\pm$ 0.03 & 0.74 $\pm$ 0.04 & 0.79 $\pm$ 0.04 & 0.33 $\pm$ 0.16 \\
XGBoost-Mordred & 0.84 $\pm$ 0.04 & 0.88 $\pm$ 0.03 & 0.89 $\pm$ 0.01 & 0.76 $\pm$ 0.06 & 0.66 $\pm$ 0.09 & 0.77 $\pm$ 0.07 & 0.25 $\pm$ 0.20 \\
RF-CheMeleonFP & 0.82 $\pm$ 0.04 & 0.86 $\pm$ 0.05 & 0.87 $\pm$ 0.01 & 0.75 $\pm$ 0.04 & 0.69 $\pm$ 0.08 & 0.78 $\pm$ 0.03 & 0.39 $\pm$ 0.06 \\
XGBoost-RDKit2d & 0.83 $\pm$ 0.01 & 0.88 $\pm$ 0.03 & 0.88 $\pm$ 0.01 & 0.74 $\pm$ 0.06 & 0.67 $\pm$ 0.10 & 0.75 $\pm$ 0.08 & 0.30 $\pm$ 0.19 \\
XGBoost-Morgan & 0.85 $\pm$ 0.02 & 0.87 $\pm$ 0.04 & 0.87 $\pm$ 0.01 & 0.73 $\pm$ 0.03 & 0.72 $\pm$ 0.05 & 0.78 $\pm$ 0.05 & 0.29 $\pm$ 0.18 \\
XGBoost-CheMeleonFP & 0.80 $\pm$ 0.06 & 0.87 $\pm$ 0.04 & 0.88 $\pm$ 0.01 & 0.75 $\pm$ 0.02 & 0.66 $\pm$ 0.07 & 0.77 $\pm$ 0.03 & 0.29 $\pm$ 0.10 \\
Chemprop GNN & 0.82 $\pm$ 0.04 & 0.86 $\pm$ 0.02 & 0.90 $\pm$ 0.01 & 0.67 $\pm$ 0.05 & 0.65 $\pm$ 0.10 & 0.72 $\pm$ 0.06 & 0.22 $\pm$ 0.06 \\
\bottomrule
\end{tabular}%
}
\end{table*}

\begin{table*}[htbp]
\centering
\caption{Polymer-solvent interaction benchmark 10-fold cross-validation $R^2$ results (mean $\pm$ standard deviation) with baseline rows. Boldface follows the rounded-mean and standard-deviation tie-breaking rule from Table~\ref{tab:engineering_combined}.}
\label{tab:fpi_interaction_r2}
\small
\setlength{\tabcolsep}{4pt}
\begin{tabular}{lc}
\toprule
Model & PolySolv \\
\midrule
TabPFN-RDKit2d & \textbf{0.93 $\pm$ 0.03} \\
TabICL-RDKit2d & \textbf{0.93 $\pm$ 0.03} \\
D-MPNN-TC~\cite{liao2025directed} & \textbf{0.93 $\pm$ 0.03} \\
CatBoost-RDKit2d & 0.90 $\pm$ 0.03 \\
CatBoost-Mordred & 0.90 $\pm$ 0.03 \\
TabPFN-Morgan & 0.88 $\pm$ 0.04 \\
TabPFN-Mordred & 0.88 $\pm$ 0.05 \\
CatBoost-CheMeleonFP & 0.88 $\pm$ 0.04 \\
XGBoost-RDKit2d & 0.88 $\pm$ 0.04 \\
TabICL-Morgan & 0.87 $\pm$ 0.04 \\
XGBoost-Mordred & 0.87 $\pm$ 0.04 \\
RF-RDKit2d & 0.85 $\pm$ 0.03 \\
CheMeleon & 0.83 $\pm$ 0.08 \\
RF-Mordred & 0.82 $\pm$ 0.05 \\
TabICL-Mordred & 0.82 $\pm$ 0.06 \\
TabPFN-CheMeleonFP & 0.81 $\pm$ 0.08 \\
XGBoost-CheMeleonFP & 0.80 $\pm$ 0.06 \\
CatBoost-Morgan & 0.79 $\pm$ 0.06 \\
RF-Morgan & 0.79 $\pm$ 0.04 \\
XGBoost-Morgan & 0.77 $\pm$ 0.07 \\
RF-CheMeleonFP & 0.74 $\pm$ 0.08 \\
Chemprop GNN & 0.72 $\pm$ 0.09 \\
TabICL-CheMeleonFP & 0.71 $\pm$ 0.10 \\
\bottomrule
\end{tabular}%
\end{table*}

\subsection{Brouwer Infinite-Dilution Activity-Coefficient Dataset}
\label{apdx:brouwer_activity_coefficients}

\begin{figure}
\centering
\includegraphics[width=0.85\linewidth]{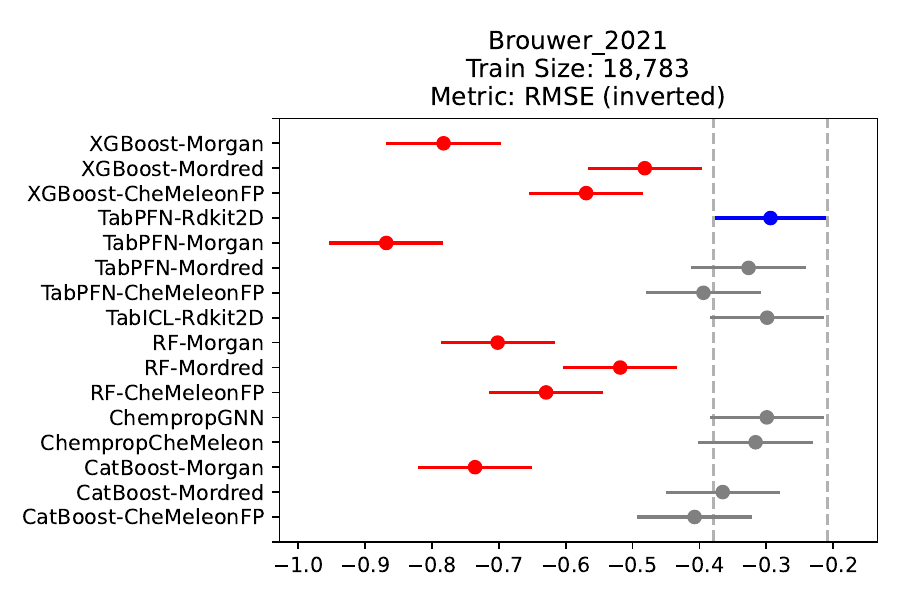}
\caption{Model performance for 10-fold cross-validation with random system-wise splits for the infinite-dilution activity-coefficient Brouwer dataset~\cite{medina2023gibbs, brouwer2021trends}. This corresponds to a \emph{non-stratified} "discrete-interpolation"~\cite{medina2023gibbs} setting. TabPFN-RDKit2d highlighted in blue is the top performer, whereas the models shown in gray are not significantly different from it according to Tukey’s honestly significant difference (HSD) test ($\alpha = 0.05$) across ten folds. Models shown in red performed significantly worse and were considered to have lost on that benchmark.}
\label{fig:hsd_brouwer_gamma}
\end{figure}

\begin{figure}
\centering
\includegraphics[width=0.85\linewidth]{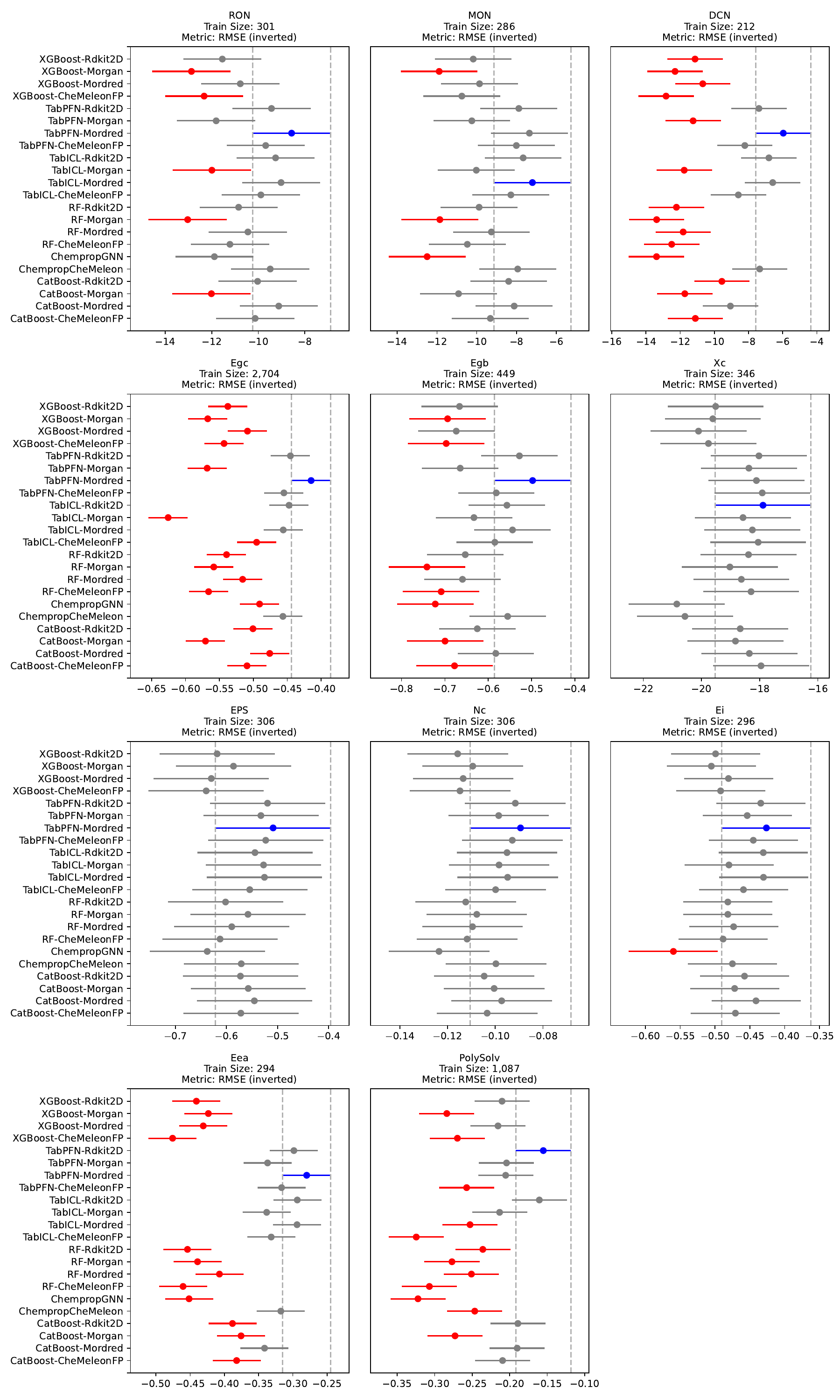}
\caption{Tukey's HSD plots ($\alpha = 0.05$) for all 11 engineering tasks.}
\label{fig:hsd_engineering}
\end{figure}

Additionally we compared a subset of all models on a larger engineering dataset, e.g., the Brouwer infinite-dilution activity-coefficient dataset~\cite{medina2023gibbs,brouwer2021trends}.
The dataset contains 20,870 measurements of the logarithmic infinite-dilution activity coefficient, $\ln \gamma^\infty$, for binary solute--solvent systems.
Each observation is defined by a solute SMILES, a solvent SMILES, and a temperature, with $\ln \gamma^\infty$ as the regression target.
After filtering to valid SMILES, the processed dataset contains 373 solutes, 349 solvents, and 6,416 distinct solute--solvent pairs, with temperatures ranging from $-23.15$ to $282.45\,^\circ\mathrm{C}$.
We used a 10-fold group split by solute--solvent pair, corresponding to a \emph{non-stratified} discrete-interpolation setting, so that all temperature measurements for a given binary pair remain in the same fold.
Each fold therefore contains 18,783 training observations and 2,087 test observations, making this benchmark substantially larger than the other engineering datasets considered in the main text.
The compact RDKit2d pair representation (435 features in total) with TabPFN performs strongest and gives the lowest mean RMSE, with RMSE $=0.294$, MAE $=0.137$, and $R^2=0.980$.
The default Chemprop GNN performs very strongly on this larger activity-coefficient benchmark and achieves strong mean 10-fold performance, with RMSE $=0.299$, MAE $=0.160$, and $R^2=0.979$.
The CheMeleon model is also strong, with an RMSE $=0.316$, but worse than the model trained from scratch.
At the same time, TabPFN-Mordred remains close to the CheMeleon result, with RMSE $=0.327$, MAE $=0.160$, and $R^2=0.975$.
A Tukey HSD plot is provided in Figure~\ref{fig:hsd_brouwer_gamma}.
The runtime comparison remains favorable for TabPFN: TabPFN-RDKit2d requires 61 s per fold and TabPFN-Mordred requires 92 s per fold on average for fitting and prediction (without feature computation), whereas Chemprop GNN requires 515 s per fold.
TabICL experiments with the larger feature sets ran out of GPU memory on a single NVIDIA H100.

\begin{table}[htbp]
\centering
\caption{Win rates and ranks across the 11 engineering datasets. Wins count models that are best or statistically indistinguishable from the best model on a task.}
\begin{tabular}{lrll}
\toprule
Model & Win Count & Win Rate (\%) & Average Rank \\
\midrule
TabPFN-Mordred & 11 & 100.0 & 2.00 \\
TabPFN-RDKit2d & 11 & 100.0 & 3.18 \\
TabICL-RDKit2d & 11 & 100.0 & 3.45 \\
TabICL-Mordred & 10 & 90.9 & 4.27 \\
TabPFN-CheMeleonFP & 10 & 90.9 & 5.73 \\
CatBoost-Mordred & 10 & 90.9 & 6.73 \\
CheMeleon & 10 & 90.9 & 8.91 \\
TabICL-CheMeleonFP & 9 & 81.8 & 9.27 \\
CatBoost-RDKit2d & 8 & 72.7 & 10.09 \\
CatBoost-CheMeleonFP & 7 & 63.6 & 10.64 \\
TabPFN-Morgan & 9 & 81.8 & 11.09 \\
TabICL-Morgan & 8 & 72.7 & 12.27 \\
RF-Mordred & 7 & 63.6 & 13.27 \\
XGBoost-Mordred & 8 & 72.7 & 14.55 \\
RF-RDKit2d & 7 & 63.6 & 14.82 \\
CatBoost-Morgan & 5 & 45.5 & 15.18 \\
XGBoost-RDKit2d & 8 & 72.7 & 15.73 \\
RF-CheMeleonFP & 6 & 54.5 & 17.27 \\
RF-Morgan & 4 & 36.4 & 17.64 \\
XGBoost-Morgan & 4 & 36.4 & 18.09 \\
XGBoost-CheMeleonFP & 5 & 45.5 & 19.09 \\
Chemprop GNN & 4 & 36.4 & 19.73 \\
\bottomrule
\end{tabular}
\label{tab:winrates_engineering}
\end{table}

\end{document}